%% file: PaperForReview.tex
\documentclass[10pt,twocolumn,letterpaper]{article}

\usepackage{iccv}
\usepackage{times}
\usepackage{epsfig}
\usepackage{graphicx}
\usepackage{amsmath}
\usepackage{amssymb}

\usepackage{amssymb}
\usepackage{booktabs}
\usepackage[export]{adjustbox} %
\usepackage{subcaption}

\usepackage{algorithm}
\usepackage[noend]{algpseudocode}
\usepackage{eqparbox}
\usepackage{arydshln}
\long\def\comment#1{}

\newcommand{\NA}[1]{--}
\DeclareMathOperator*{\argmax}{arg\,max}

\usepackage{array}
\newcolumntype{H}{>{\setbox0=\hbox\bgroup}c<{\egroup}@{\hspace*{-\tabcolsep}}}

\usepackage[pagebackref=true,breaklinks=true,letterpaper=true,colorlinks,bookmarks=false]{hyperref}

\iccvfinalcopy %

\def\iccvPaperID{9941} %

\ificcvfinal\fi

\begin{document}

\title{
EgoAdapt: A multi-stream evaluation study of adaptation \\ to real-world egocentric user video
}

\author{
Matthias De Lange\textsuperscript{$\dagger,\ddagger$}
\\
Michael Louis Iuzzolino\textsuperscript{$\dagger$}
\and
Hamid Eghbal-zadeh\textsuperscript{$\dagger$}\\
Franziska Meier\textsuperscript{$\dagger$}
\and
Reuben Tan\textsuperscript{$\dagger$}\\
Karl Ridgeway\textsuperscript{$\dagger$}
\and \and \textsuperscript{$\dagger$}Meta AI \quad \textsuperscript{$\ddagger$}KU Leuven\\
}

\maketitle
\ificcvfinal\thispagestyle{empty}\fi

\begin{abstract}
In egocentric action recognition a single population model is typically trained and subsequently embodied on a head-mounted device, such as an augmented reality headset.
While this model remains static for new users and environments, we introduce an adaptive paradigm of two phases, where after pretraining a population model, the model adapts on-device and online to the user's experience.
This setting is highly challenging due to the change from population to user domain and the distribution shifts in the user's data stream.
Coping with the latter \mbox{in-stream} distribution shifts is the focus of continual learning, where progress has been rooted in controlled benchmarks but challenges faced in real-world applications often remain unaddressed.
We introduce \textbf{EgoAdapt}, a benchmark for real-world egocentric action recognition that facilitates our two-phased adaptive paradigm, and real-world challenges naturally occur in the egocentric video streams from Ego4d, such as long-tailed action distributions and large-scale classification over 2740 actions. 
We introduce an evaluation framework that directly exploits the user's data stream with new metrics to measure the adaptation gain over the population model, online generalization, and hindsight performance. In contrast to single-stream evaluation in existing works, our framework proposes a meta-evaluation that aggregates the results from 50 independent user streams.
We provide an extensive empirical study for finetuning and experience replay.\footnote{Code is made publicly available at \scriptsize{\url{https://github.com/facebookresearch/EgocentricUserAdaptation}}
}
\end{abstract}

\section{Introduction}

\begin{figure}[t]
\centering
\includegraphics[clip,trim={0.3cm 0.1cm 2cm 0.2cm},width=1\linewidth]{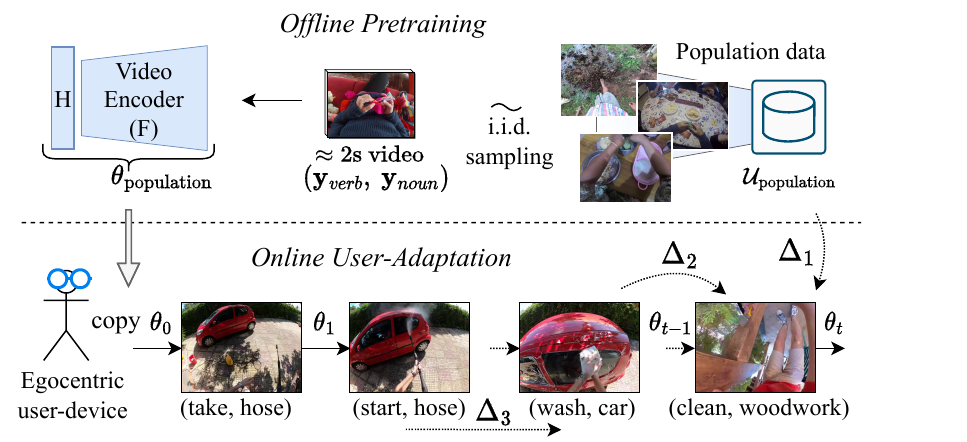} 
\caption{\label{fig:setup_main_figure}
\textbf{EgoAdapt} focuses on the gain of on-device adaption to the user (from \emph{top} to \emph{bottom} row), and online learning with natural distribution shifts (\emph{bottom} row). 
While only a single user stream is depicted, EgoAdapt enables a meta-evaluation of 50 independent user streams.
Distribution shifts occur once from population to user ($\Delta_1$), and continually during adaptation in both the input domain ($\Delta_2$) and action distribution ($\Delta_3$).
In the \emph{top} row, a static model $f_{\theta_\text{population}}$ is learned from egocentric video over a vast population of users ($\mathcal{U}_\text{population}$). 
The \emph{bottom} row depicts subsequent adaptation to the user's experience, after model initialization with $f_{\theta_\text{population}}$.
All user video is obtained from Ego4d~\cite{ego4d}.
\looseness=-1
}
\end{figure}

One of the cornerstones to improving human-computer interaction is for machine-learning systems to understand or predict human behavior~\cite{HCIintelligent}.
A head-mounted device, such as an augmented reality headset, enables a first-person viewpoint from the user, 
where recognizing the user's actions is key to building such improved understanding.
Current egocentric action recognition models aim for generalization to new users through training on videos from a vast and diverse population of users~\cite{Feichtenhofer_2019_ICCV,Fan_2021_ICCV}.
However, this population model remains unchanged on the user device, disregarding factors that are highly prone to change over time such as the surrounding environment or user behavior and preferences.
\looseness=-1

The main focus of this work is to improve generalization for a specific user by adapting online to new experiences over time. 
The setting we consider is specifically challenging due to the combination of three desiderata.
First, Learning a user-specific expert model should result in improvement over the initial user-agnostic population model.
Second, as user data becomes only gradually available over time, the expert model should adapt online to the user's new experience.
Third, while retaining user-specific knowledge is desirable, learning from new data with distribution shifts may result in catastrophic forgetting of previous knowledge~\cite{french1999catastrophic}.

An obstacle to evaluate these three desiderata is that in real-world data streams, clear held-out evaluation tasks are typically unavailable.
Therefore, we propose an evaluation framework that directly exploits the user's data stream to measure our novel metrics for \emph{Adaptation Gain} over the population model, the online generalization, and the hindsight performance.
Using this evaluation framework, we aim to empirically study the learning behavior of standard stochastic gradient descent  and experience replay in continual learning.

Existing continual learning benchmarks are often artificially created from static datasets~\cite{lopez2017gradient,aljundi2019online,De_Lange_2021_ICCV,chrysakis20a},
and their focus is confined to adaption on a single stream~\cite{Wang_2021_ICCV,lin2021clear,cai2021online}. 
Furthermore, to date, no continual learning benchmark exists for egocentric action recognition.
To this end, our empirical study focuses
on an extremely challenging real-world benchmark for continual learning that introduces many aspects often neglected in existing benchmarks. We summarize their limitations in the following.
\looseness=-1
\begin{enumerate}
    \item Real-world data distributions may have limited and application-specific guarantees. This may result in an imbalance between classes, dependencies that result in correlated data streams, large output spaces, and natural re-occurrences of classes in the data stream. 
    \item Standard practice of analyzing learning behavior on a single data stream may introduce biased results. This is especially undesirable as continual learning methodologies are desired to be stream-agnostic, while the data streams at deployment may be prone to high variability.
    \looseness=-1
    \item Existing works focus mainly on image classification, neglecting the context in the video stream.
\end{enumerate}

We propose the egocentric action recognition benchmark \textbf{EgoAdapt}, addressing all three limitations with challenging real-world data, multiple independent user streams, and focusing on video context for action recognition.
Additionally, EgoAdapt enables evaluation for our three desiderata by means of two controlled phases, first pretraining over a population of users, followed by a phase of online adaptation over user-specific data streams. 
In the second phase, in contrast to existing works evaluating a single stream, EgoAdapt entails video from Ego4d~\cite{ego4d} for 50 independent real-world  user streams from the egocentric perspective, allowing a meta-evaluation over the streams with our proposed evaluation framework.
The variety and scale in this real-world benchmark make it particularly interesting for our study, spanning 53 different scenarios with 2740 unique actions over 77 hours of annotated video.

Our study finds that personalization offers significant improvement for users over the population model even with simple online finetuning, while adapting the features or revisiting samples with ER greatly ameliorates forgetting without losing online generalization performance.
Our transfer study between user models indicates the models become true experts of the user stream, with significant improvement over the population model but trading off generalization to other user streams.
\looseness=-1

\section{Related Benchmarks}
\noindent\textbf{Continual Learning benchmarks} are typically
constructed by manually grouping subsets of static datasets in a sequence of tasks~\cite{surveyCLtpamiMatthias,van2019three}, for example Rotated-MNIST~\cite{lopez2017gradient} or Core50~\cite{lomonaco2017core50}.
Such task-based continual learning has been explored for non-local user adaptation in the cloud~\cite{Lange_2020_CVPR}.
The task boundaries allow constructing held-out evaluation sets a priori to measure per-task performance, which is typically infeasible for real-world agents that are oblivious to plausible future tasks.
Recent works propose real-world datasets without task boundaries and alternative evaluation schemes for autonomous driving~\cite{verwimp2022clad}, and long-term concept evolution in YFCC100M~\cite{YFCC100M} for image classification~\cite{lin2021clear} and geolocalization~\cite{cai2021online}. 
Wanderlust~\cite{Wang_2021_ICCV} considers frame-based egocentric object detection spanning 18 hours of video in outdoor scenes over nine months of a graduate student's life.
In contrast to existing benchmarks, EgoAdapt enables video-based prediction from the egocentric perspective, focuses on large-scale action classification over 2740 actions, and provides 50 independent task-agnostic user streams in the real world instead of a single stream.
\looseness=-1

\textbf{Egocentric action recognition benchmarks} are often scripted, predetermining which actions a participant should record~\cite{fathi2012learning,damen2014you,de2009guide,pirsiavash2012detecting}.
As this work focuses on natural real-world distribution shifts, to date two large-scale egocentric datasets entail unscripted video.
EPIC-KITCHENS-100~\cite{EpicKitchensDamen2022RESCALING} contains 100 hours of video but is limited to users in a kitchen environment.
In contrast, the Ego4d~\cite{ego4d} forecasting benchmark comprises 110 hours of video in 53 different scenarios in everyday activities.
Ego4d stands out in terms of diversity and scale with data collected by 7 worldwide universities in different countries, 7 varieties of head-mounted recording devices, and 406 participants.

\section{Online Egocentric User-Adaptation}
Here we formalize the setup, followed by the EgoAdapt benchmark details in Section~\ref{sec:online_action_recog_benchmark}, as summarized in Figure~\ref{fig:setup_main_figure}.
The user-adaptation setup consists of two phases.
First, a user-agnostic population model is optimized over a population of users.
Second, the local user device starts with the population model but adapts the model the user's experience over time.
In pretraining the population model, no resource constraints are imposed, and typically large amounts of data and computational resources are available.
In contrast, for continual learning on the local user device, the data is processed in a streaming fashion, storing only the most recent observed data for processing, with an additional fixed memory capacity for continual learning methods.

\noindent\textbf{Formalization.} 
The data stream of user $u$ is defined as $S_u = \left\{ ({\bf x}_t,{\bf y}_t) \right\}^{|S_u|-1}_{t=0}$ with size $|S_u|$, and sample $({\bf x}_t,{\bf y}_t)$ at time step $t$ consisting of video-input ${\bf x}_t$ and supervision signal ${\bf y}_t$.
As is common practice in online continual deep learning, sample $({\bf x}_t,{\bf y}_t)$ at time step $t$ may concern a small batch rather than a single sample~\cite{De_Lange_2021_ICCV}.
The user's predictive model ${\bf \Tilde{y}}_t = f_{\theta_t}\left( {\bf x}_t \right)$ is parameterized by $\theta_t$ before updating with $({\bf x}_t,{\bf y}_t)$, and is initialized with the population model $\theta_0 \leftarrow \theta_{\text{population}}$.
Parameters are updated by optimizing a loss function $\mathcal{L}({\bf \Tilde{y}}_t, \ {\bf y}_t)$, given prediction ${\bf \Tilde{y}}_t$ and ground truth ${\bf y}_t$, denoted as $\mathcal{L}_t$ in short.
Note that we omit the model's user-subscript to avoid clutter. 
We assume users are part of mutually exclusive sets, with users $u\in\mathcal{U}_{population}$ included to pretrain the population model, $\mathcal{U}_{train}$ to select hyperparameters, and $\mathcal{U}_{test}$ as a held-out evaluation set.
Note that our extensive ablation study deliberately focuses on the 10 user streams in $\mathcal{U}_{train}$, mainly for computational feasibility and consistency in the
study, as for example examining the relations between users is quadratic (requiring $1.6$k entries for $\mathcal{U}_{test}$, and only $100$ for $\mathcal{U}_{train}$).
The general setup %
is depicted in Figure~\ref{fig:setup_main_figure}.

\subsection{An online action-recognition benchmark}\label{sec:online_action_recog_benchmark}
To construct a real-world benchmark for user-adaptation, we consider three key factors.
First, the data should be collected over time and exhibit natural distribution shifts. 
Second, we require video meta-data indicating the user, with a sufficient number of users and data per user. 
Third, the dataset should contain users with diverse geographical and demographic backgrounds.
The Ego4d forecasting benchmark~\cite{ego4d} fulfills all requirements.
We consider the combined data of the publicly available Ego4d training and validation splits for action-based forecasting, resulting in a total of 77 hours of annotated video.
User streams are constructed by grouping the video data per participant in Ego4d.
\looseness=-1

\begin{figure}[!h]

\centering
\caption{\label{fig:egoAdapt_main_stats}
(\emph{top}) \textbf{User splits} indicated in color, with users ordered on video length in minutes.
(\emph{center})
\textbf{Action distribution ($\text{P}_\text{action}$) shift} from $\mathcal{U}_{population}$ to $\mathcal{U}_{test}$ with actions ordered on frequency in $\mathcal{U}_{population}$.
(\emph{bottom})
\textbf{Per-user and average $\text{CDF}_\text{action}$} respectively indicated as colored lines and black markers, with actions per user ordered from high to low frequency.
}%

\begin{subfigure}{1\linewidth}
    \centering
    \includegraphics[clip,trim={0.2cm 0.2cm 0cm 0.2cm},width=1\linewidth]{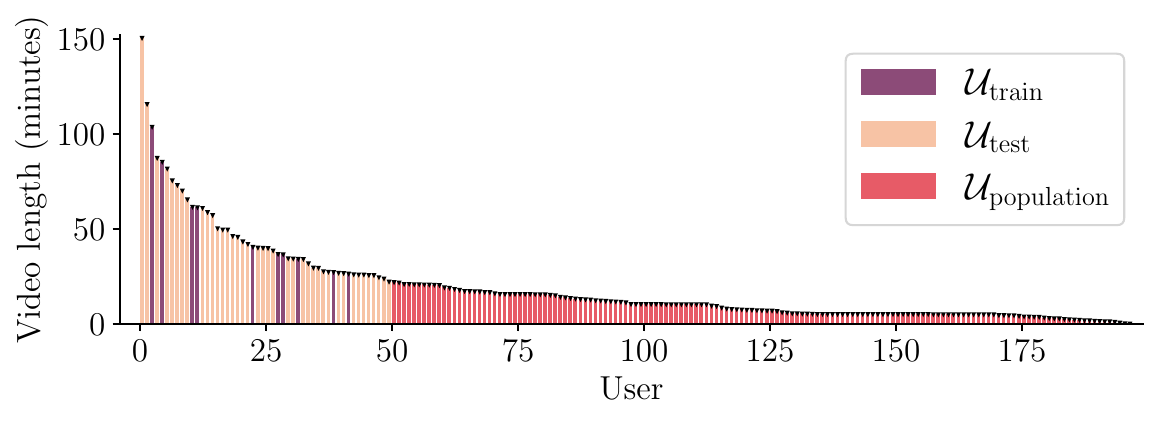} %
\end{subfigure}
\begin{subfigure}{1\linewidth}
    \centering
    \includegraphics[clip,trim={0.2cm 0cm 0.2cm 0.2cm},width=1\linewidth]{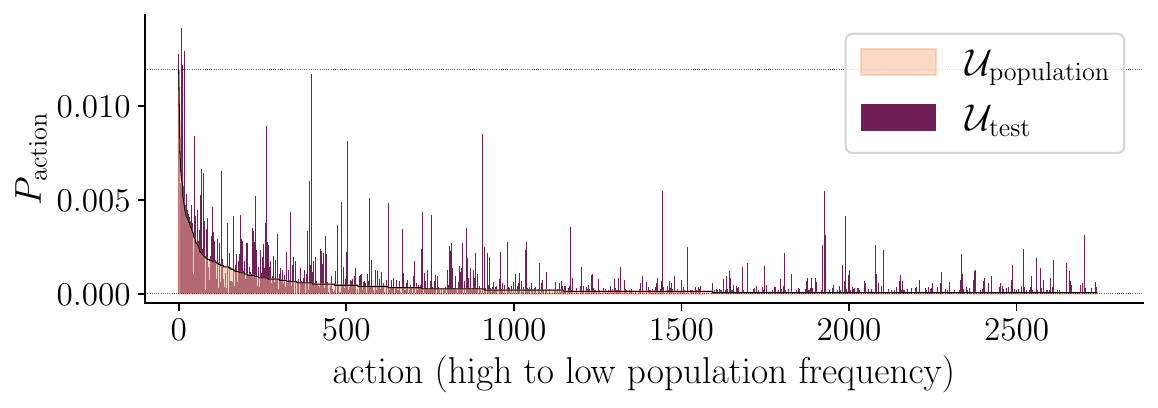}
\end{subfigure}
\begin{subfigure}{1\linewidth}
    \centering
    \includegraphics[clip,trim={0.2cm 0cm 0.2cm 0.2cm},width=1\linewidth]{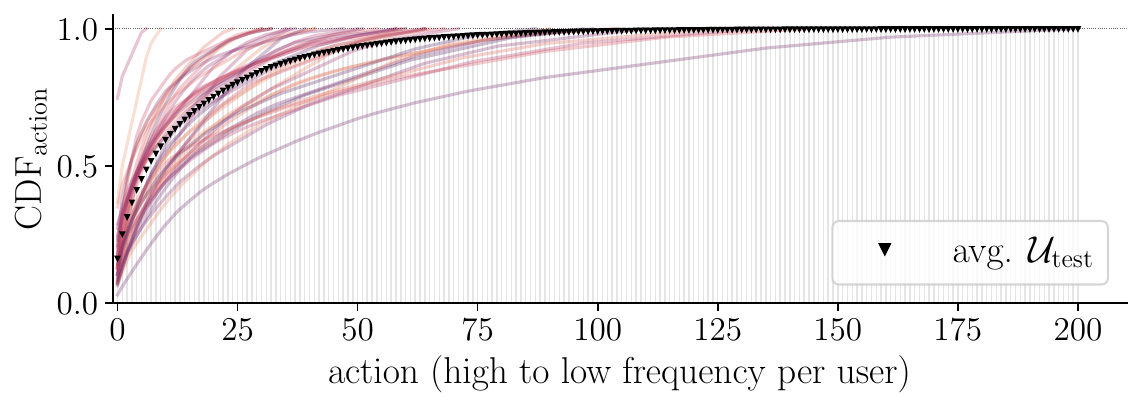} 
\end{subfigure}
\vspace{-0.5cm}
\end{figure}

\noindent\textbf{User splits} are shown in Figure~\ref{fig:egoAdapt_main_stats}(\emph{top}) with the total video length per user.
As users require sufficient data to analyze adaptation, we select the 50 users with the largest amount of video data. These are then randomly subdivided in 10 users in $\mathcal{U}_{\text{train}}$ (9 hours) and 40 in $\mathcal{U}_{\text{test}}$ (31 hours). 
We exploit the remaining participant data (15 hours) and additionally consider video without participant meta-data (22 hours) as single-video users for $\mathcal{U}_{\text{population}}$.
Figure~\ref{fig:egoAdapt_main_stats}(\emph{center}) indicates the significant shift for the action distribution $\text{P}_\text{action}$ from $\mathcal{U}_{\text{population}}$ to $\mathcal{U}_{\text{test}}$.

\noindent\textbf{Long-tailed Action Recognition.} 
Given a input clip ${\bf x}_t$ of $2.1$ seconds at time step $t$, the network comprising a video encoder and action classifier, should predict the correct action ${\bf y}_t= \left( {\bf y}_{verb, t}, {\bf y}_{noun, t} \right)$, consisting of a verb ${\bf y}_{verb, t}$ and noun ${\bf y}_{noun, t}$.
The distributions over actions, verbs, and nouns in the user streams are long-tailed. Figure~\ref{fig:egoAdapt_main_stats}(\emph{bottom}) shows the cumulative action distribution function ($\text{CDF}_\text{action}$) for all users in $\mathcal{U}_{\text{test}}$, obtained by normalizing action-histograms, sorted from high to low frequency. The $\text{CDF}_\text{action}$ per user indicates a large variety in the total number of actions per user stream, but all users exhibit a long-tailed action distribution.
The results for $\mathcal{U}_{\text{train}}$ and verb and noun CDFs can be found in Appendix.
\looseness=-1

\noindent\textbf{Setup.} Following action recognition literature, the nouns and verbs are predicted by two independent classifiers~\cite{ego4d}.
To maintain comparability of results, we consider the standard Ego4d SlowFast~\cite{Feichtenhofer_2019_ICCV} video encoder based on Resnet101~\cite{He2015}. At each time step we consider a mini-batch of 4 consecutive samples.
In preprocessing of the streams we omit video segments without annotations and give precedent to earlier actions to the intersection of overlapping action segments. 
EgoAdapt focuses on domain adaptation from population to user domain, hence considers in the user streams only the 107 verbs and 384 nouns observed during pretraining.
Further details can be found in Appendix and provided code.

\section{User-Adaptation Metrics}
To learn a user model online, we identify three main factors to quantify: (1) model performance compared to the population model; (2) model generalization for unseen samples in the stream; (3) performance retention on the observed part of the stream.
To this end, we propose two metrics that both directly compare the improvement over the population model $f_{\theta_0}$, called the \textbf{Adaptation Gain (AG)}.
Given a base metric $\phi$ for which higher is better, the AG is defined as:
\looseness=-1
\begin{equation}
    \text{AG}_{\theta_t} \left( {\bf x}_i,{\bf y}_i \right) = \phi( {\bf y}_i, f_{\theta_t}({\bf x}_i)) - \phi( {\bf y}_i, f_{\theta_0}({\bf x}_i))
\end{equation}
for the user-adapted model $f_{\theta_t}$ at time step $t$.
In the following, we use by default the class-balanced or macro-average accuracy (\textbf{ACC}) as base metric $\phi$, or denote with subscript $\mathcal{L}$ when reporting the loss objective over samples. 
Note that the class-balancing in ACC re-weighs from long-tailed to uniform class distribution.

The \textbf{Online Adaptation Gain ($\text{OAG}$)} measures the AG of currently observed samples at time step $t$ before updating $\theta_t$. Accumulating the AG over these unseen samples in stream $S$ gives an indication of online generalization.
\looseness=-1
\begin{equation}
    \text{OAG}_t \left( S \right) = \sum_{k=0}^t \text{AG}_{\theta_k} \left( {\bf x}_k, {\bf y}_k \right)
\end{equation}
Second, besides adapting to the distribution shifts in the stream, it is desirable for the learner to maintain the previously acquired knowledge in the stream.
Therefore, we propose the \textbf{Hindsight Adaptation Gain (HAG)} measuring the AG over the full observed subset of the stream $S$ on the current model $\theta_t$.
\begin{equation}
    \text{HAG}_t \left( S \right) =  \sum_{k=0}^t \text{AG}_{\theta_t} \left( {\bf x}_k, {\bf y}_k \right)
\end{equation}

\noindent\textbf{Stream aggregation metrics.}
To quantify the OAG and HAG over multiple user streams of various lengths, aggregation is required.
We adopt a uniform prior over the users and normalize user streams to the per-sample average.
Per user $u \in \mathcal{U}$ the final adaptation performance is considered at the end of learning stream user stream $S_u$ with $t=|S_u|$.
This results in the following metrics:
\begin{align}
    \overline{\text{OAG}} &= \sum_{u \in \mathcal{U}} |S_u|^{-1} \text{OAG}_{|S_u|} \left( S_u \right) \\
    \overline{\text{HAG}} &= \sum_{u \in \mathcal{U}} |S_u|^{-1} \text{HAG}_{|S_u|} \left( S_u \right) 
\end{align}
Additionally, we denote the action, verb, and noun metrics by means of subscript as in $\overline{\text{OAG}}_\text{action}$.

\section{Empirical study}

\subsection{Action non-stationarity analysis}
\label{sec:non-stat_analysis}
As the actions in real-world video streams are naturally highly correlated over time, we first aim to quantify the span of temporal consistency for the action, verbs, and nouns in a user stream.
To this end, we introduce the \emph{Label-Window Predictor} (LWP),
storing a window of the $W$ most recent observed labels to predict the most frequent one. The accuracy metric of the LWP quantifies the temporal consistency as it indicates how well previous samples can predict the subsequent one.
Table~\ref{tab:label_window_predictor} reports the average class-balanced $\overline{\text{ACC}}$ over user streams, confirming the strong correlation of actions, verbs, and nouns with $W=1$.
However, as the window $W$ increases and more context is considered, the highest frequency label in the window deteriorates as predictor. This indicates the natural non-stationarity of the action distribution over longer time spans (large $W$), while locally strongly correlated over time (small $W$).
\looseness=-1

\begin{table}[!h]
\caption{\label{tab:label_window_predictor}
The \emph{Label-window predictor} (LWP) predicts the most frequent label in a window of size $W$. Results are reported as class-balanced accuracy with mean ($\pm \text{SE}$) over users in $\mathcal{U}_{\text{train}}$.
\looseness=-1
}
\centering
\resizebox{0.8\linewidth}{!}{%
\begin{tabular}{rrrr}
\toprule
 \multicolumn{1}{r}{$W$} & $\overline{\text{ACC}}_{\text{action}}$ & $\overline{\text{ACC}}_{\text{verb}}$ & $\overline{\text{ACC}}_{\text{noun}}$ \\
\midrule
                               1 &                                     $40.9\pm2.2$ &                                   $43.5\pm3.3$ &                                   $54.1\pm1.8$ \\
                               4 &                                     $14.8\pm1.2$ &                                   $21.8\pm2.2$ &                                   $28.6\pm2.1$ \\
                              32 &                                      $4.3\pm0.9$ &                                    $8.7\pm1.1$ &                                   $10.8\pm1.4$ \\
                        unlimited &                                      $2.9\pm0.7$ &                                    $7.6\pm0.9$ &                                    $6.8\pm1.3$ \\
\bottomrule
\end{tabular}
}
\end{table}

\subsection{User-Adaptation with online finetuning}
\begin{figure}[!h]
\centering
\caption{\label{fig:SGD_per_user_online_OAG}
\textbf{Finetuning improves online over the population model}. 
Reports $\text{OAG}_{\mathcal{L},\text{action}}$, the OAG for the cumulative action-loss (y-axis), over time step iterations per user stream (x-axis), for the 10 users in $\mathcal{U}_{\text{train}}$ (colored lines).
}%
\includegraphics[clip,trim={0.2cm 0cm 0.6cm 0.2cm},width=1\linewidth]{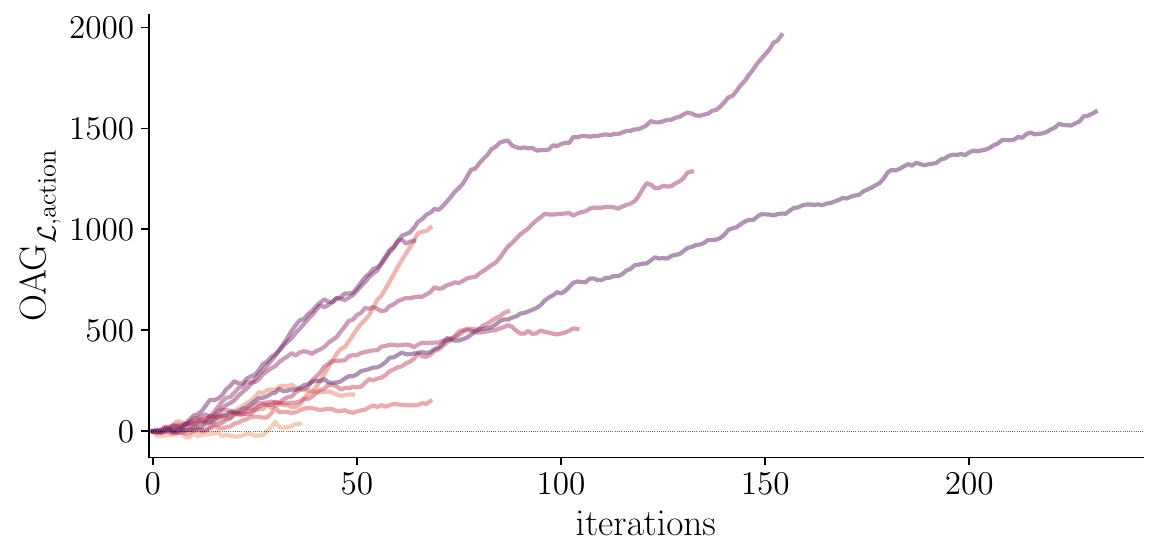} 
\end{figure}

Online finetuning uses plain stochastic gradient descent (SGD) to learn in a single pass from the temporally ordered mini-batches in a user stream. 
In this and the following experiments, we follow common practice in online continual learning by processing small mini-batches~\cite{aljundi2019online,aljundi2019gradient,De_Lange_2021_ICCV}, here set to 4 consecutive video clips of $2.1$ seconds.
Finetuning typically results in worst-case performance in continual learning, as it is highly prone to catastrophic forgetting~\cite{surveyCLtpamiMatthias,van2019three}.
However, Figure~\ref{fig:SGD_per_user_online_OAG} shows for all users in $\mathcal{U}_{\text{train}}$ online generalization improvement over the population model, reporting the cumulative action-loss compared to the population model, i.e. the $\text{OAG}^{}_{\mathcal{L},\text{action}}$ per user. 
A single user initially performs slightly worse than the population model, but recovers near 30 iterations.
Averaged over users in $\mathcal{U}_{\text{train}}$ ($\pm \text{SE}$), the following table shows that the online generalization $\overline{\text{OAG}}_{\text{action}}$ is larger than the hindsight performance $\overline{\text{HAG}}_{\text{action}}$. 
\begin{table}[!h]
\vspace{-0.2cm}
\resizebox{\linewidth}{!}{%
\begin{tabular}{lll|lll}
 $\overline{\text{OAG}}_{\text{action}}$ & $\overline{\text{OAG}}_{\text{verb}}$ & $\overline{\text{OAG}}_{\text{noun}}$ & $\overline{\text{HAG}}_{\text{action}}$ & $\overline{\text{HAG}}_{\text{verb}}$ & $\overline{\text{HAG}}_{\text{noun}}$ \\
\midrule
                                    $4.9\pm1.2$ &                                        $5.5\pm1.6$ &                                        $8.9\pm1.5$ &                                        $2.6\pm0.8$ &                                        $3.6\pm1.2$ &                                        $4.8\pm1.7$ \\
\end{tabular}
}
\vspace{-0.4cm}
\end{table}

\noindent\emph{This is surprising as this indicates that performance is better for unseen samples, than for samples that have been observed before.}
This behavior might be caused by the high plasticity of SGD in highly correlated data streams: adapting quickly to the most recent batch is likely to perform better for the next batch, with the cost of forgetting previous knowledge. 

In the following, we investigate the effects of adapting the features and head of the model, and how multiple updates on a single batch may further improve results.
Additionally, given the strong temporal correlation of the actions, we hypothesized using momentum would accelerate adaptation.
We empirically found this is not the case, and perform an analysis of gradient direction in finetuning that indicates subsequent gradients are often interfering. The momentum results and gradient analysis can be found in Appendix due to space constraints.

\subsubsection{Learning user-specific features}\label{sec:feat_vs_classifier}
We can disentangle the predictive function $f_\theta \equiv F_{\theta_F} \circ H_{\theta_H}$ as the composition of two subsequent operations: extracting the features with function $F$, followed by generating a prediction from the features with classifier head $H$.
Based on a learning rate grid search for $\overline{\text{OAG}}$, Table~\ref{tab:frozen_feat} reports results for optimizing the full model ($F \circ H$), compared to $F$ or $H$ only.
Only optimizing the feature extractor $F$ with a fixed classifier from the population model results in a significant improvement
with positive $\overline{\text{OAG}}$ and $\overline{\text{HAG}}$. This indicates the merits of adapting the features to the user.
For optimizing only the classifier $H$ large improvement in $\overline{\text{OAG}}$ can be observed. This is to be expected due to adaptation to a limited number of actions per user compared to the 2740 actions in the population model (see Figure~\ref{fig:egoAdapt_main_stats}c). 
Optimizing only the classifier $H$ results in small $\overline{\text{OAG}}$ improvement over optimizing the full model ($F \circ H$), as also observable for our final benchmark results for the 40 users in $\mathcal{U}_\text{test}$ in Table~\ref{tab:final_test_user_table}.
Interestingly, for hindsight performance in  Table~\ref{tab:frozen_feat}, optimizing the full model results in at least $1.7$, $3.1$, and $4.9$ absolute increase over optimizing only $H$ in  $\overline{\text{HAG}}$ for actions, verbs, and nouns.
We further analyze this observation in the following.

\begin{figure}[!h]
\caption{
\textbf{Linear noun-classifier analysis} comparing learning the full model $F \circ H$ or the classifier $H$ only.
Per user the final classifier weight and bias $L2$-norms are compared to the initial population model.
The per-user delta-distribution is averaged over users in $\mathcal{U}_{\text{train}}$, shown with shaded $SE$. Decreases w.r.t. the population model are displayed as negative.
Nouns are ordered based on mass in the average noun distribution $P_{\text{label}}$ over streams (gray area). 
}
\label{fig:classifier_only_weight_biases_analysis}
\centering
        \begin{subfigure}{0.5\linewidth}
      \centering
        \caption{weight norm delta}%
        \includegraphics[clip,trim={0.2cm 0.2cm 0cm 0.2cm},width=1\textwidth]{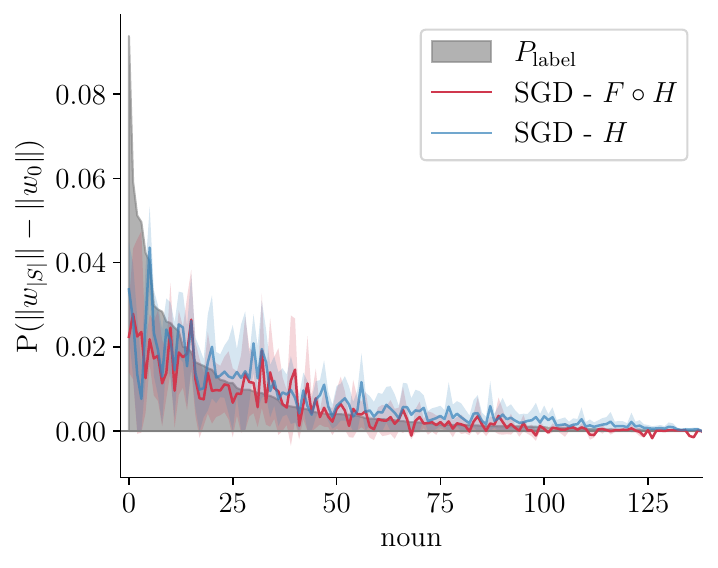} %
    \end{subfigure}%
        \begin{subfigure}{0.5\linewidth}
      \centering
              \caption{bias norm delta}%
        \includegraphics[clip,trim={0.2cm 0.2cm 0cm 0.2cm},width=1\textwidth]{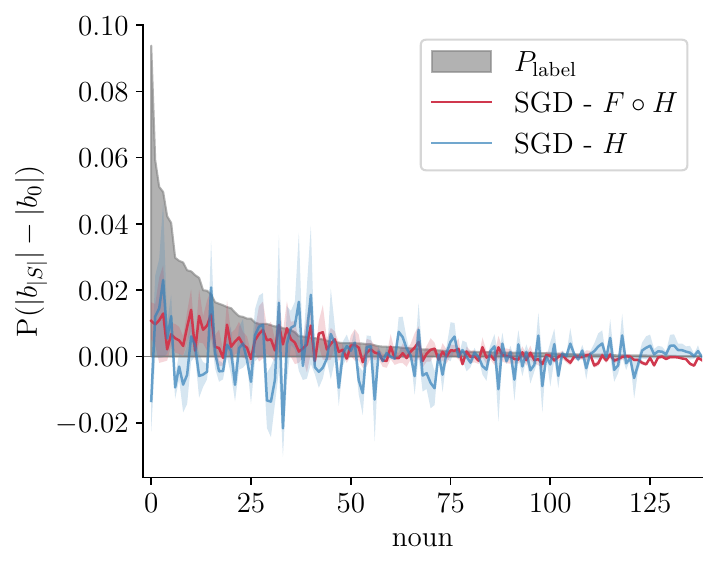} %
    \end{subfigure}%
    \vspace{-0.3cm}
\end{figure}

Given a feature $F({\bf x})$, $H$ is defined by two independent linear classifiers for verbs and nouns.
The classifier $H(F({\bf x})) = \argmax_y  F({\bf x}) {\bf w}_{y} + b_y$ can increase the score for the correct class $y_c$ in two ways: increase the magnitude of the corresponding weight vector ${\bf w}_{y_c}$, or increase the bias $b_{y_c}$.
Figure~\ref{fig:classifier_only_weight_biases_analysis} shows the noun-classifier changes in  weight and bias magnitude in hindsight for the final user model compared to the population model.
The weights and biases are ordered based on the total frequency over user streams.
Learning the full model exhibits a trend of following the noun-frequency in the streams. 
However, learning only the classifier shows large decreases in bias for several high-frequency nouns. 
This finding indicates how learning the full model retains better hindsight performance over the streams.
\looseness=-1

\begin{table}[!h]
\caption{\textbf{Feature and classifier adaptation} after initialization with the population model, optimizing only the feature extractor ($F$), the classifier head ($H$), or both ($F \circ H$). 
Reported as mean ($\pm \text{SE}$) over user streams in $\mathcal{U}_{\text{train}}$. 
}
\label{tab:frozen_feat}
\resizebox{\linewidth}{!}{%
\begin{tabular}{lrrr|rrr}
\toprule
  optimize & $\overline{\text{OAG}}_{\text{action}}$ & $\overline{\text{OAG}}_{\text{verb}}$ & \multicolumn{1}{c}{$\overline{\text{OAG}}_{\text{noun}}$} & \multicolumn{1}{c}{$\overline{\text{HAG}}_{\text{action}}$} & $\overline{\text{HAG}}_{\text{verb}}$ & $\overline{\text{HAG}}_{\text{noun}}$ \\
\midrule
                     $F$ &                                        $1.8\pm0.6$ &                                        $1.4\pm0.7$ &                                        $5.0\pm1.4$ &                                        $0.9\pm0.3$ &                                        $0.3\pm0.5$ &                                       $-0.1\pm0.6$ \\
$H$ &                                        $5.3\pm0.9$ &                                        $7.3\pm0.9$ &                                       $12.0\pm2.0$ &                                        $0.9\pm0.3$ &                                        $0.5\pm0.4$ &                                       $-0.2\pm0.7$ \\
$F \circ H$ &                                        $4.9\pm1.2$ &                                        $5.5\pm1.6$ &                                        $8.9\pm1.5$ &                                        $2.6\pm0.8$ &                                        $3.6\pm1.2$ &                                        $4.8\pm1.7$ \\

\bottomrule
\end{tabular}
}
\end{table}

\subsubsection{Multiple updates for a single batch}
\label{sec:multi_iter}
In online learning each sample is only observed once. However, as is common practice in online continual learning, the same batch can be reprocessed to accommodate better gradient-based learning~\cite{aljundi2019online,De_Lange_2021_ICCV}.
We apply the same principle in Table~\ref{tab:SGD_multi_iter}, showing both increased online generalization ($\overline{\text{OAG}}_{\text{action}}$) and hindsight performance ($\overline{\text{HAG}}_{\text{action}}$) up to 10 updates with the same mini-batch.
Additionally, 
we report the 
batched version of the LWP in Section~\ref{sec:non-stat_analysis} ($\text{LWP}_{\text{B}}$), updating the window $W$ only after predicting for the entire current batch $B_t$ rather than per instance. This reference baseline gives an indication of performance when perfectly fitting the current batch labels. Nonetheless, learning for 10 iterations significantly outperforms $\text{LWP}_{\text{B}}$.
To gain further insights in the online generalization, we additionally split the $\overline{\text{OAG}}_{\text{action}}$ on all user stream data, into the same metric on correlated ($\overline{\text{OAG}}^\text{cor.}_{\text{action}}$) and decorrelated ($\overline{\text{OAG}}^\text{decor.}_{\text{action}}$) data. The split is considered correlated if the same action is observed at the previous time step and decorrelated on action transitions, respectively resulting in a $76/24\%$ split 
of $\mathcal{U}_{\text{train}}$.
Table~\ref{tab:SGD_multi_iter} shows insignificant changes for $\overline{\text{OAG}}^\text{decor.}_{\text{action}}$, while the correlated data gains significant improvements with multiple updates per batch. 
Appendix reports up to 50 iterations, with no significant effect above 10 updates.

\begin{table}[!h]
\caption{
\textbf{Finetuning multiple updates per batch} increases hindsight performance ($\overline{\text{HAG}}_{\text{action}}$) and online generalization ($\overline{\text{OAG}}_{\text{action}}$), further decomposed in decorrelated ($\overline{\text{OAG}}^\text{decor.}_{\text{action}}$) and correlated data ($\overline{\text{OAG}}^\text{cor.}_{\text{action}}$). 
Reports mean ($\pm \text{SE}$) over $\mathcal{U}_{\text{train}}$. 
}
\label{tab:SGD_multi_iter}
\centering
\resizebox{0.9\linewidth}{!}{%
\begin{tabular}{rrr|rr}
\toprule
\multicolumn{1}{c}{updates} & \multicolumn{1}{c}{$\overline{\text{OAG}}_{\text{action}}$} & \multicolumn{1}{c}{$\overline{\text{HAG}}_{\text{action}}$} & \multicolumn{1}{c}{$\overline{\text{OAG}}^\text{decor.}_{\text{action}}$} & \multicolumn{1}{c}{$\overline{\text{OAG}}^\text{cor.}_{\text{action}}$} \\
\midrule
$\emph{LWP}_{\text{B}}$ & $6.0\pm1.3$ & $0.4\pm0.4$ & $1.5\pm0.8$ &                                                  $8.4\pm1.7$ \\
\addlinespace
1 &                                        $4.9\pm1.2$ &                                        $2.6\pm0.8$ &                                        $2.8\pm1.0$ &                                        $6.3\pm1.5$ \\
                         2 &                                        $6.2\pm1.1$ &                                        $2.6\pm0.5$ &                                        $2.7\pm1.1$ &                                        $8.4\pm1.4$ \\
                         3 &                                        $7.6\pm1.3$ &                                        $3.4\pm0.7$ &                                        $2.8\pm1.1$ &                                       $10.3\pm1.5$ \\
                         5 &                                        $7.9\pm1.4$ &                                        $4.3\pm1.1$ &                                        $2.5\pm1.1$ &                                       $11.4\pm1.7$ \\
                        10 &                                        $8.7\pm1.3$ &                                        $4.7\pm1.3$ &                                        $3.0\pm1.1$ &                                       $12.1\pm1.5$ \\
\bottomrule
\end{tabular}
}
\end{table}

\subsection{User-Adaptation with Experience Replay}

Previous results showed online finetuning to significantly improve over the population model. However, online generalization to unseen samples excels over the performance of learned samples in hindsight.
This is undesirable as we aim for a trade-off in quick adaptation to the current samples, while retaining this knowledge as learning continues.
A standard strategy in continual learning is the use of a replay memory $\mathcal{M}$, where $M$ observed samples are stored and later revisited~\cite{Verwimp_2021_ICCV,chaudhry2019continual,aljundi2019online}.
We examine three policies on deciding which samples are stored in $\mathcal{M}$ while using random retrieval from the memory to add a batch of identical size to the current mini-batch for learning.
Firstly, we consider a first-in-first-out (\textbf{FIFO}) storage policy, keeping only the $M$ most recent observed samples.
The second storage policy uses reservoir sampling~\cite{vitter1985random} where once $\mathcal{M}$ is full, each sample at time step $t$ has probability $M/t$ to be stored with random replacement ($\textbf{Reservoir}$).
However, the sampling is class-independent, which may result in $\mathcal{M}$ mainly containing samples from the stream's majority classes.
This limitation is addressed by class-balanced reservoir sampling (CBRS)~\cite{chrysakis20a}, dividing the memory over observed classes, each maintained by the use of reservoir sampling.
A shortcoming of CBRS is the assumption of a larger memory size than the number of observed actions. This is not the case in our setup, as many actions occur in a stream while the memory-demanding video samples constrain the memory size.
Therefore, we propose a hybrid solution of the class-balanced reservoir sampling (\textbf{Hybrid-CBRS}), that falls back to reservoir sampling once the number of observed classes is greater than or equal to the memory size $M$.
The method is described in Algorithm~\ref{algor:CBRS_adapted} in Appendix.
\looseness=-1

\begin{figure*}[!ht]
\begin{minipage}{\textwidth}
  \begin{minipage}[t]{0.6\textwidth}
    \centering
        \captionof{table}{\label{tab:final_test_user_table}
        \textbf{EgoAdapt test user results} of the 40 user streams in $\mathcal{U}_{\text{test}}$, reported as mean ($\pm SE$).
        Bold results indicate best online user-adaptation results with the same capacity, excluding \emph{SGD-i.i.d.} and \emph{ER-Full} baselines.
        }
        \vspace{-0.3cm}
    \resizebox{\textwidth}{!}{%
    \input{tables/final_test_results_tabular_only.tex}
    }
  \end{minipage}
  \hfill
  \begin{minipage}[t]{0.38\textwidth}
    \centering
        \captionof{figure}{
        \label{fig:transfer_eval}
        \textbf{User transfer matrix} for users in $\mathcal{U}_{\text{train}}$.
        Rows represent user-adapted models after learning the user stream.
        Columns evaluate a row's user model on the various user streams.
        Reports the loss in hindsight compared to the population model as $\text{HAG}_{\mathcal{L},\text{action}}$.
        \looseness=-1
    }
    \vspace{0.4cm}
    \includegraphics[clip,trim={0.4cm 0.2cm 2cm 1.2cm},width=\textwidth]{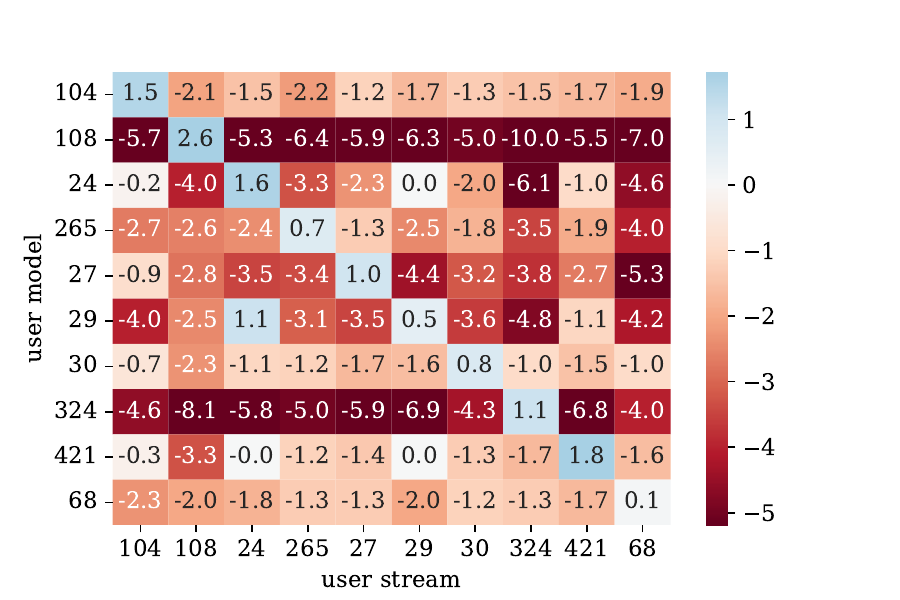} %
  \end{minipage}
\end{minipage}
\end{figure*}

\noindent\textbf{Results.} Table~\ref{tab:replay} compares the storage strategies for a range of memory sizes $M$ and compares to baselines storing all samples (\textbf{ER-Full}) or none at all (SGD).
All ER results perform consistently similar to SGD in terms of online generalization. 
However, hindsight performance is significantly improved  even for a memory size of only two batches (8 samples).
This is expected as ER repeatedly optimizes for samples observed in the stream, but interestingly this has no significant decrease in the online generalization.
Both Reservoir and Hybrid-CBRS outperform the FIFO strategy in hindsight as FIFO revisits only the recent correlated samples.
The results in Table~\ref{tab:replay} report over users in $\mathcal{U}_{\text{train}}$, and looking ahead to our final results with $\mathcal{U}_{\text{test}}$ in Table~\ref{tab:final_test_user_table}, we observe Hybrid-CBRS to significantly outperform both Reservoir and ER-Full in hindsight.

\noindent\textbf{ER feature adaptation.}
To get insights in the improved hindsight performance of ER in comparison with SGD, we conduct an analysis of the feature quality produced by feature extractor $F$.
To this end, after learning from the user stream, we assess the representation quality using linear probing~\cite{Chen_2021_ICCV,he2020momentum}. Due to the lack of held-out data in the real-world user streams, we train and evaluate this ideal hindsight classifier re-using the user's data stream.
If ER mainly affects adaptation of the classifier, both the user-adapted models for ER and SGD should result in similar performance.
We compare SGD with the best-performing ER using Hybrid-CBRS and memory size 64, and retrain the classifier for 10 epochs.
Table~\ref{tab:replay_classifier_retrain} shows that ER attains significantly better memorization performance than SGD, indicating improved feature adaptation contributes to the increased hindsight performance.

\begin{table}[!h]
\caption{
\textbf{Experience Replay (ER)} for three  storage policies and memory sizes $M$.
\emph{ER-Full} stores all samples, and \emph{SGD} stores none.
Reported as mean ($\pm \text{SE}$) over users in $\mathcal{U}_{\text{train}}$. 
\looseness=-1
}
\label{tab:replay}
\centering
\resizebox{0.8\linewidth}{!}{%
\begin{tabular}{lrll}
\toprule
          Storage Policy &  $M$ & $\overline{\text{OAG}}_{\text{action}}$ & $\overline{\text{HAG}}_{\text{action}}$ \\
\midrule
\textbf{FIFO} &                                    8 &                                        $4.5\pm1.0$ &                                        $8.6\pm2.0$ \\
                     &                                   64 &                                        $3.7\pm0.9$ &                                       $15.7\pm2.6$ \\
                     &                                  128 &                                        $4.0\pm1.0$ &                                       $18.7\pm2.3$ \\
          \addlinespace$\textbf{Reservoir}$ &                                    8 &                                        $3.5\pm1.0$ &                                       $13.6\pm1.7$ \\
           &                                   64 &                                        $3.9\pm0.9$ &                                       $24.8\pm3.2$ \\
           &                                  128 &                                        $3.9\pm0.8$ &                                       $24.0\pm2.5$ \\
       \addlinespace$\textbf{Hybrid-CBRS}$                  &        8 &                                        $3.9\pm1.0$ &                                       $15.6\pm2.5$ \\
                            &      64 &                                        $4.6\pm0.9$ &                                       $29.7\pm4.7$ \\
                            &     128 &                                        $4.1\pm0.9$ &                                       $25.1\pm4.1$ \\                     
\midrule\textbf{ER - Full}                     &                              $\infty$ &                                        $3.8\pm0.9$ &                                       $23.3\pm2.5$ \\
\textbf{SGD}    &     0 &                                                        $4.9\pm1.2$ &                                                                         $2.6\pm0.8$  \\

\bottomrule
\end{tabular}
}
\end{table}

\begin{table}[!h]
\caption{
\textbf{ER feature adaptation} is measured by evaluating the stream classification performance (ACC) after retraining the final user model classifiers.
Reported as mean ($\pm \text{SE}$) over users in $\mathcal{U}_{\text{train}}$ for ER with Hybrid-CBRS storage policy ($M=64$) and SGD. 
}
\label{tab:replay_classifier_retrain}
\centering
\resizebox{0.8\linewidth}{!}{%
\begin{tabular}{lrrr}
\toprule
Method & \multicolumn{1}{c}{$\overline{\text{ACC}}_{\text{action}}$} & \multicolumn{1}{c}{$\overline{\text{ACC}}_{\text{verb}}$} & \multicolumn{1}{c}{$\overline{\text{ACC}}_{\text{noun}}$}  \\
\midrule
   \textbf{SGD} &                                                $19.6\pm2.7$ &                                              $25.3\pm3.8$ &                                              $29.7\pm4.0$ \\
   \textbf{ER} &$46.9\pm3.8$ &                                              $48.9\pm3.7$ &                                              $52.5\pm4.4$ \\
\bottomrule
\end{tabular}
}
\end{table}

\subsection{User-adaptation and forgetting}
\begin{figure}[!h]
\caption{
\textbf{Re-exposure forgetting (RF)} of all 782 re-occurrences of 270 actions in $\mathcal{U}_{\text{train}}$ streams.
Samples are grouped in 10 bins after log-scaling of the re-exposure iterations for better spread.
Reporting mean ($\pm SE$) per bin for ER with Hybrid-CBRS storage policy ($M=64$) and SGD.
}
\label{fig:forgetting_analysis}
\centering
    \begin{subfigure}{1\linewidth}
      \centering
        \includegraphics[clip,trim={0.4cm 0.6cm 0.6cm 0.5cm},width=1\textwidth]{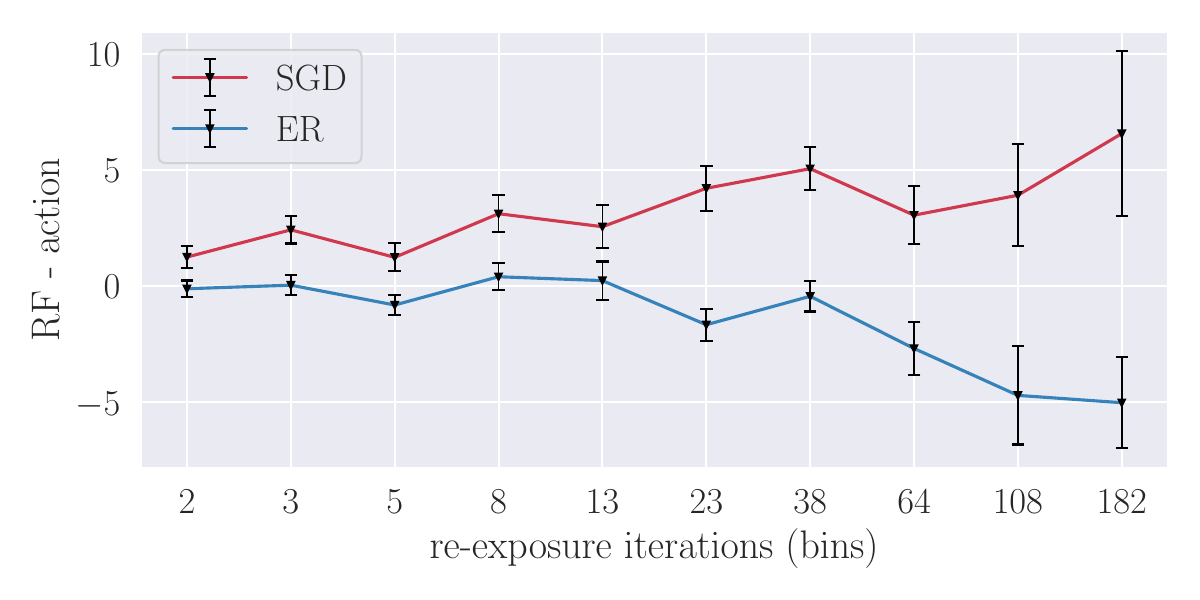} %
    \end{subfigure}
\end{figure}

Catastrophic forgetting due to non-stationarity in the user stream
 is problematic for personalization as besides quick adaptation, it is desirable to maintain good performance on the observed stream.
In continual learning, the performance loss or \emph{forgetting} is measured on held-out datasets from clearly distinct tasks~\cite{surveyCLtpamiMatthias}.
In real-world data streams with natural distribution shifts, it remains unclear how to measure forgetting.
Therefore, we propose a label-conditional evaluation of forgetting without requiring clearly defined evaluation tasks. 
To this end, we measure how performance of an action is affected before it naturally re-occurs in the data stream.
Between the two occurrences of action ${\bf y}_t$, the learning of other actions may interfere and induce forgetting of ${\bf y}_t$.
Hence, two models should be compared: 
first, $f_{\theta_{t+1}}$ after updating on an occurrence at time step $t$ of ${\bf y}_t$; second, $f_{\theta_e}$ with $e>t+1$ just before update of the next instance $({\bf x}_e,{\bf y}_e)$ with ${\bf y}_e = {\bf y}_t$.
To measure the delta on the exact same data, all samples in the stream before and including time step $t$ with label ${\bf y}_t$ are considered.
This results in the \textbf{re-exposure forgetting (RF)} for observed action $y_t$ at time step $t$:
\begin{equation}
    \text{RF} =  |S^{{\bf y}_t}_{0:t}|^{-1} \sum_{({\bf x}_i,{\bf y}_i) \in S^{{\bf y}_t}_{0:t}} \mathcal{L}( {\bf y}_i, f_{\theta_e}({\bf x}_i)) - \mathcal{L}( {\bf y}_i, f_{\theta_{t+1}}({\bf x}_i))
\end{equation}
with $S^{{\bf y}_t}_{0:t}$ the stream subset up to and including time step $t$ for samples with label ${\bf y}_t$. The \emph{average-RF}  averages over all re-exposures to summarize all considered user streams.
For all re-occurrences in $\mathcal{U}_{\text{train}}$, Figure~\ref{fig:forgetting_analysis} shows the RF in function of the number of iterations before re-exposure. For visualization, the re-exposure iterations are first log-scaled, then grouped in 10 bins, reporting mean and SE per bin.
The RF increases for a larger number of iterations between two exposures for SGD, resulting in an average-RF of $2.6 \pm 0.26$. 
However, ER shows negative RF for a larger number of iterations between exposures with an average-RF of $-0.63 \pm 0.20$.
This indicates the efficacy of revisiting the data for a larger number of iterations in ER, rather than inducing larger forgetting as in SGD.

\subsection{User transfer study}
To validate the knowledge transfer between user models, we construct a \textbf{user transfer matrix} for all users in $\mathcal{U}_{\text{train}}$, where each user expert model (row) is evaluated on all user data streams (columns). 
We report the $\overline{\text{HAG}}_{\mathcal{L},\text{action}}$, as the learning loss $\mathcal{L}$ allows further insights beyond zero accuracy, and the metric compares directly to the population model's performance on the stream. %
The matrix in Figure~\ref{fig:transfer_eval} confirms the efficacy of user adaptation with the highest adaptation gain attained on the diagonal.
For the off-diagonal entries, user models in general perform worse than the population model with negative $\overline{\text{HAG}}_{\mathcal{L},\text{action}}$. 
This indicates that user-adaptation results in user-expert models while sacrificing generalization to other users.
We observe two remarkable results in the transfer matrix.
First, users $324$ and $108$ are expert models that have poor transfer to any other users.
Second, user stream $24$ results in better performance for the model of user $29$ compared to the population model.
In Appendix (Figure~\ref{APDX:fig:transfer_eval:IOU}), we report the intersection-over-union (IOU) for the actions, verbs, and nouns between the users, indicating similar actions for users 24 and 29, with $23\%$ overlap for the action domain, and $69\%$ for verbs,  $35\%$ for nouns.
\looseness=-1

\section{Final benchmark results on test users}

Table~\ref{tab:final_test_user_table} summarizes our findings averaged over the 40 test user streams in $\mathcal{U}_\text{test}$, reporting the class-balanced accuracy ($\overline{\text{ACC}}_\text{action}$) for online and hindsight performance on actions, verbs, and nouns. Note that in contrast to adaptation gain in our empirical study, absolute results are reported to enable easy comparison for follow-up works, independent of the pretraining performance.
The \textbf{Random} classifier indicates classification difficulty, resulting in $0.9$, and $0.3$ accuracy for classifying 107 verbs and 384 nouns respectively, with $2.4e^{-3}$ accuracy for all verb-noun combinations for actions.
The population model pretrained with $\mathcal{U}_\text{population}$ indicates effective pretraining by attaining $1.2$ $\overline{\text{ACC}}_{\text{action}}$, significantly outperforming \emph{Random}.

\noindent\textbf{Label-window predictor.} From our experiments in Section~\ref{sec:multi_iter}, we consider both 1 and 10 updates per batch in the stream, reporting the $\text{LWP}_\text{B}$ baseline for perfectly fitting the current batch classes for predicting the next. Table~\ref{tab:final_test_user_table} shows our conclusions hold for $\mathcal{U}_\text{test}$ with multiple SGD gradient updates significantly outperforming $\text{LWP}_\text{B}$.

\noindent\textbf{Finetuning the classifier.} We compare finetuning the classifier head only (\textbf{SGD-head only}) with the full model (SGD), resulting in $1\%$ and $0.4\%$ improvement in online $\overline{\text{ACC}}_{\text{action}}$ for 1 and 10 updates per batch. However, in hindsight finetuning the head only results in only a small improvement over the population model.

\noindent\textbf{Breaking correlation.} Subsequently, to measure the influence of strongly correlated user streams, \textbf{SGD-i.i.d.} breaks the correlation by shuffling the user stream,  resulting in an identical and independently sampled distribution (i.i.d).  Comparing \emph{SGD} with \emph{SGD-i.i.d.} shows a decrease in online generalization, whereas hindsight $\overline{\text{ACC}}_{\text{action}}$ exhibits an increase from $5\%$ to $15\%$. This might indicate that the temporal correlation induces significant forgetting and hence deteriorates memorization of the stream.

\noindent\textbf{Experience Replay (ER)} with the various storage strategies indicates similar online generalization performance to SGD, while significantly improving the hindsight performance. Especially updating 10 times per batch is beneficial when allowing resampling from the memory.
Noteably, our Hybrid-CBRS storage strategy outperforms and approaches storing all samples (\textbf{ER-Full}) for respectively 1 and 10 updates per batch.
\looseness=-1

\section{Conclusion}
In this work, we proposed \emph{EgoAdapt}, a new egocentric action recognition benchmark for online continual learning on real-world user-specific video streams.
EgoAdapt aims to move beyond the static deployment of a pretrained population model on user devices by adapting to the user's experience. 
The 50 real-world user streams based on Ego4d enabled a meta-evaluation over the streams, and 
we introduced \emph{Adaptation Gain} metrics to directly measure improvement over the population model.
Our comprehensive empirical study indicated significant online adaptation gain with simple finetuning, while adapting the features and revisiting data with experience replay (ER) allow better retaining previous knowledge without sacrificing generalization.
With this work, we hope to foster continual learning towards real-world applications and inspire subsequent benchmarks to tackle additional open challenges such as open-world learning of the actions and reducing supervision in user streams.
\looseness=-1

\input{output.bbl}
\clearpage
\appendix
\section*{Appendix}

\section{EgoAdapt: Reproducibility details}

\noindent\textbf{Codebase.}
The Pytorch-based~\cite{paszke2017automatic} codebase uses Pytorch Lightning~\cite{Falcon_PyTorch_Lightning_2019} and enables a high level of concurrency, enabling concurrently processing multiple independent user streams on multiple devices.
It is made publicly available for reproducibility.

\noindent\textbf{Model architecture.}
We use SlowFast~\cite{Feichtenhofer_2019_ICCV} as video encoder, mapping input video to a single compressed feature representation. The temporal resolution in the \emph{Fast} pathway is 4 times  higher than the \emph{Slow} pathway ($\alpha$), while the \emph{Fast} pathway uses only 1/8 of the channels ($\beta$). The base network of SlowFast is Resnet101~\cite{He2015}.
The video representation is then used to classify actions using two independent, linear verb and noun classifiers, following~\cite{ego4d}.

\noindent\textbf{Data processing.}
All data is obtained from the publicly available Ego4d~\cite{ego4d} dataset, specifically from the forecasting benchmark.
We consistently use a batch size of $4$ video samples for online learning on the user streams. The original Ego4d data is 30FPS, from which 32 frames are sampled with sampling rate 2 to obtain a single video sample of $2.1$ seconds.
Frames in the video are scaled to $256$ pixels based on the shorter side, and center cropped on $224$ pixels. No random transforms are used to make sure hindsight performance metrics represent memorization.

\noindent\textbf{Pretraining a population model.}
To maintain a reference model in terms of performance, we follow the pretraining protocol in Ego4d~\cite{ego4d}, initializing from a Kinetics-400~\cite{kinetics400} pretrained model to avoid starting from scratch, and subsequently training on Ego4d.
The learning rate is $1e^{-4}$, and we omit Ego4d's linear warmup phase. 
Computational capacity remains similar as the 30 epochs of pretraining on full Ego4d are converted to 46 epochs for $\mathcal{U}_\text{population}$, both with the approximately the same number of training iterations.
The user streams in $\mathcal{U}_\text{population}$ are used as validation set to select the model with lowest $\mathcal{L}_{\text{action}}$, resulting in the model at 45 epochs.

\section{Experiment details and additional results}

\subsection{Online finetuning and multiple updates per batch} %

In our experiments, online Finetuning uses vanilla stochastic gradient descent (SGD) for training on the user streams. We perform a learning rate grid search $\eta \in \{0.1,0.01,0.001\}$ and select the best run based on highest class-balanced $\overline{\text{ACC}}_{\text{action}}$.

\begin{figure}[t]
\caption{
\textbf{Multiple updates per batch} for SGD.  (a) Reports online generalization ($\overline{\text{OAG}}_{\text{action}}$) and hindsight performance ($\overline{\text{HAG}}_{\text{action}}$). (b) Decomposes data for online generalization in decorrelated ($\overline{\text{OAG}}^\text{decor.}_{\text{action}}$) and correlated data ($\overline{\text{OAG}}^\text{cor.}_{\text{action}}$). 
Reported as mean ($\pm \text{SE}$) over user streams. 
}
\label{APDX:fig:multi_updates_over10}
\centering
        \begin{subfigure}{1\linewidth}
      \centering
          \caption{$\overline{\text{OAG}}_{\text{action}}$ and $\overline{\text{HAG}}_{\text{action}}$}
        \includegraphics[clip,trim={0.2cm 0cm 0.2cm 0.2cm},width=1\textwidth]{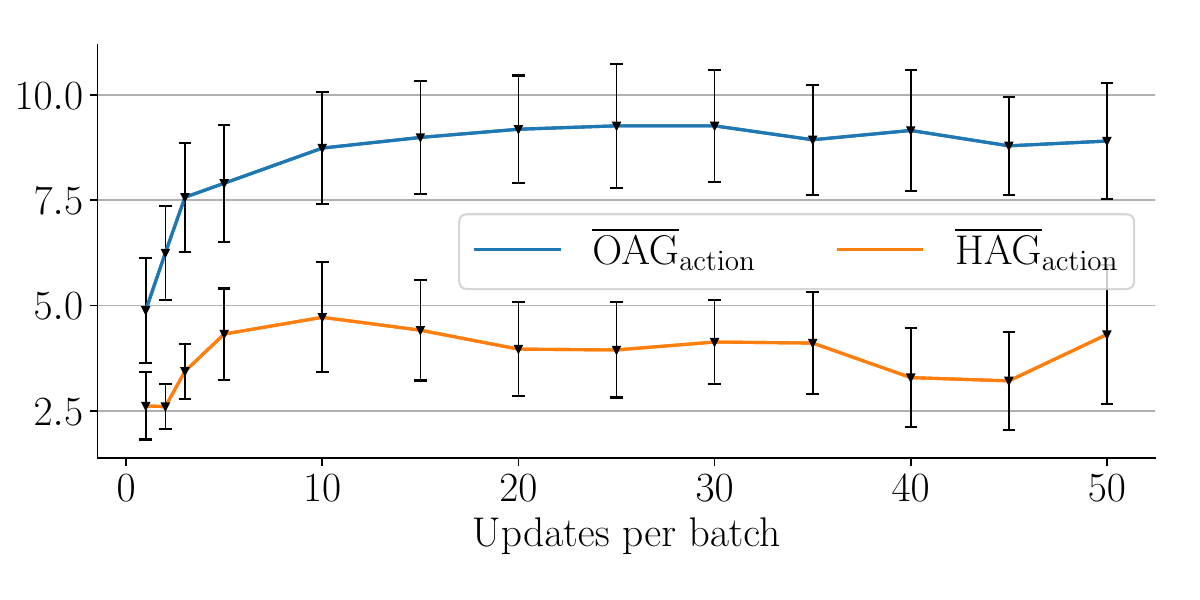} %
    \end{subfigure}
    \vspace{10pt} 
        \begin{subfigure}{1\linewidth}
      \centering
        \caption{$\overline{\text{OAG}}^\text{decor.}_{\text{action}}$ and $\overline{\text{OAG}}^\text{cor.}_{\text{action}}$}%
        \includegraphics[clip,trim={0.2cm 0.2cm 0.2cm 0.2cm},width=1\textwidth]{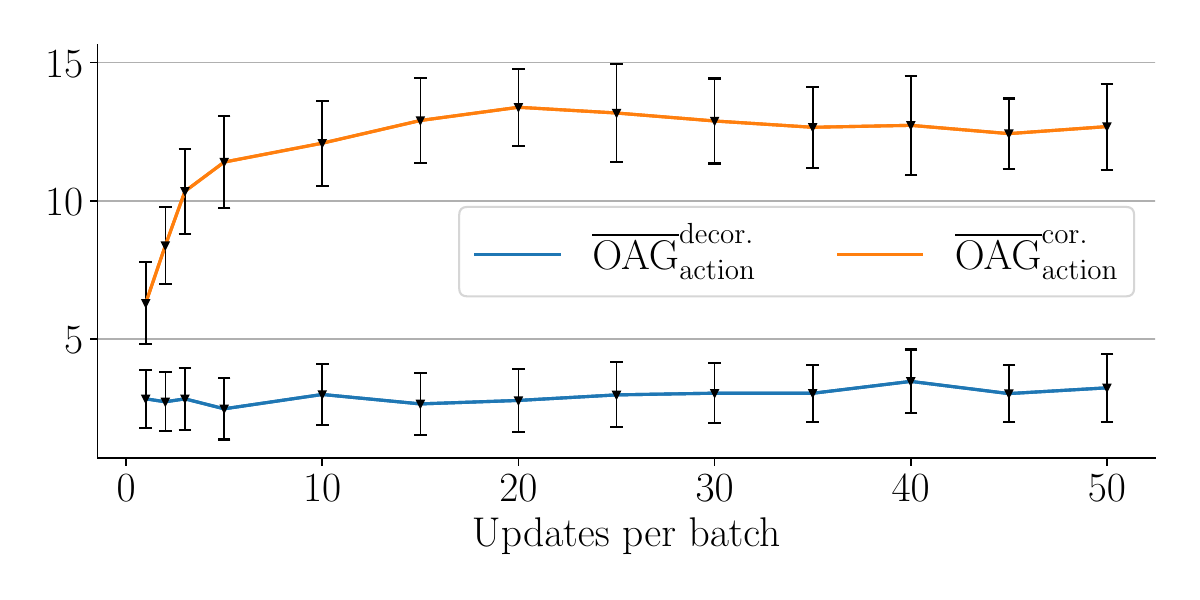} 
    \end{subfigure}%

\end{figure}

In the experiments with multiple updates per batch, the main paper reports results up to 10 updates per batch. We investigate higher number of updates per batch in steps of 5 updates, up to 50 updates. Figure~\ref{APDX:fig:multi_updates_over10} indicates no significant change in both online generalization and hindsight performance by further increasing the number of updates.

\subsection{Momentum for user-adaptation} %
Given the strong temporal correlation of the actions, we hypothesize finetuning might significantly benefit from the use of momentum to accelerate adaptation.
We examine both Nesterov-momentum and regular momentum.

\noindent\textbf{Setup.} We compare SGD for a range of momentum strengths $\rho \in \{0, 0.3,0.6,0.9\}$ for both Nesterov and regular momentum, and perform a learning rate grid search for the full network  $\eta \in \{0.1,0.01,0.001\}$, selecting the run on highest class-balanced $\overline{\text{ACC}}_{\text{action}}$.
With $\eta = 0.01$ consistently having the best results over the momentum strengths, this learning rate is used in the ablation with momentum on the classifier or feature extractor only.

\noindent\textbf{Results.}
The results for momentum on the full model can be found in Table~\ref{APDX:tab:momentum_full_model}, with Table~\ref{APDX:tab:momentum_head_classifier} reporting specifically for classifier and feature extractor only. 
In the following discussions, we focus on Nesterov-momentum, as we consistently find it to have better online generalization over plain momentum.
Similar to Section~\ref{sec:feat_vs_classifier} we consider the influence of classifier and feature extractor separately.
Table~\ref{tab:momentum_nesterov} shows for a range momentum strengths $\rho\in \{0.3, \ 0.6, \ 0.9\}$ the online generalization $\overline{\text{OAG}}$ for classifier and feature extractor both separately and combined.
In the three cases we find decreasing $\overline{\text{OAG}}$ with increasing $\rho$, indicating momentum has not the desired effect of accelerating adaptation.
\looseness=-1

\begin{table}[!h]
\caption{
\textbf{Nesterov-momentum for user-adaptation} shows declining online generalization $\overline{\text{OAG}}_{\text{action}}$ for increasing momentum strength $\rho$, reported as mean ($\pm \text{SE}$) over $\mathcal{U}_{\text{train}}$.
\looseness=-1
}
\label{tab:momentum_nesterov}
\centering
\resizebox{0.7\linewidth}{!}{%
\begin{tabular}{rlllllll}
\toprule
 \multicolumn{1}{c}{$\rho$} &  $\rho_{\text{head+feat}}$ & $\rho_{\text{head}}$ & $\rho_{\text{feat}}$\\
\midrule
                        0.0 &                                        $4.9\pm1.2$ &  \NA{} & \NA{} \\
                        0.3 &                                       $4.5\pm1.1$ & $4.2\pm1.1$ & $4.8\pm1.2$  \\
                        0.6 &                                        $3.4\pm1.1$ & $3.4\pm1.1$ & $4.3\pm1.1$ \\
                        0.9 &                                       $1.7\pm0.8$ & $2.0\pm0.8$  & $3.7\pm1.2$ \\
\bottomrule
\end{tabular}
}
\end{table}

\begin{figure}[!h]
\centering
\caption{\label{fig:SGD_grad_analysis}
\textbf{Finetuning gradient alignment} analyzed by cosine-similarity of batch gradient $g_t$ at time step $t$ with the previous gradients $k$ update steps before $t$ in the learning trajectory.
Reports mean ($\pm$ SE) over users in $\mathcal{U}_{\text{train}}$.
\looseness=-1
}%
\includegraphics[clip,trim={0.6cm 0.8cm 0.8cm 0.8cm},width=1\linewidth]{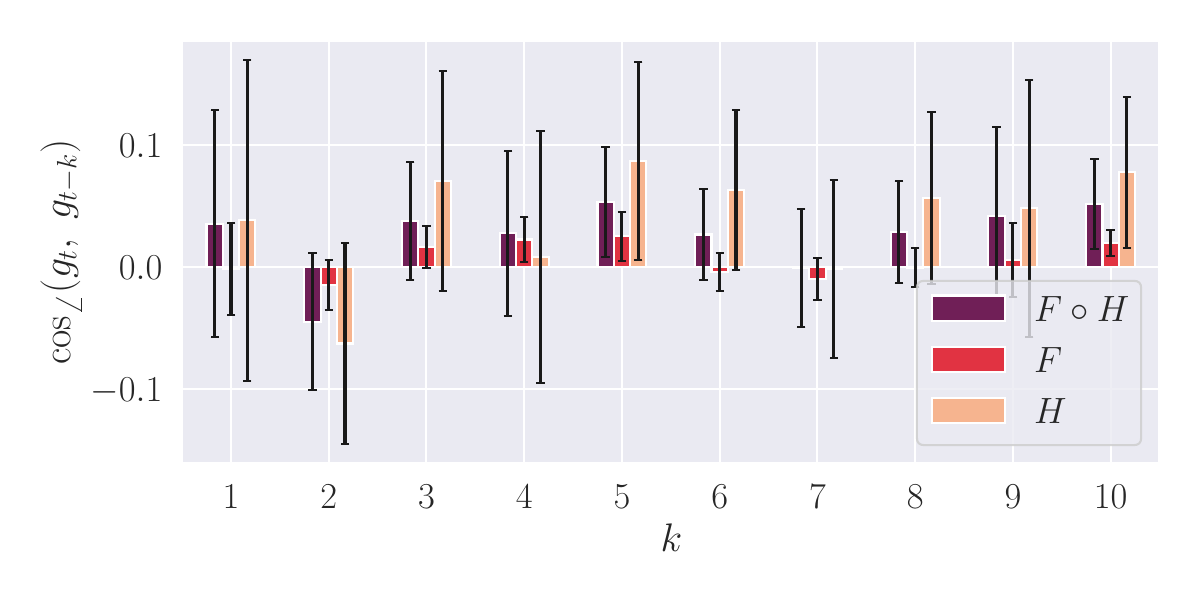} 
\vspace{-0.8cm}
\end{figure}

\begin{table}[b]
\caption{\textbf{Momentum and Nesterov-momentum results} for a grid search over learning rate $\eta$ and momentum strength $\rho$ for the full model.
Reports mean ($\pm \text{SE}$) over users in $\mathcal{U}_{\text{train}}$.
}
\label{APDX:tab:momentum_full_model}
\centering
\resizebox{\linewidth}{!}{%
\begin{tabular}{rlllllll}
\toprule
 \multicolumn{1}{c}{$\rho$} & \multicolumn{1}{c}{$\eta$} & \multicolumn{1}{c}{$\overline{\text{OAG}}_{\text{action}}$} & \multicolumn{1}{c}{$\overline{\text{OAG}}_{\text{verb}}$} & \multicolumn{1}{c}{$\overline{\text{OAG}}_{\text{noun}}$} & \multicolumn{1}{c}{$\overline{\text{HAG}}_{\text{action}}$} & \multicolumn{1}{c}{$\overline{\text{HAG}}_{\text{verb}}$} & \multicolumn{1}{c}{$\overline{\text{HAG}}_{\text{noun}}$} \\
\midrule
\multicolumn{3}{l}{\textbf{SGD}}\\
                        0.0 &                      0.001 &                                        $1.1\pm0.5$ &                                        $1.1\pm0.5$ &                                        $2.0\pm1.1$ &                                        $3.5\pm0.8$ &                                        $4.6\pm1.2$ &                                        $6.0\pm1.1$ \\
                        0.0 &                       0.01 &                                        $4.9\pm1.2$ &                                        $5.5\pm1.6$ &                                        $8.9\pm1.5$ &                                        $2.6\pm0.8$ &                                        $3.6\pm1.2$ &                                        $4.8\pm1.7$ \\
                        0.0 &                        0.1 &                                        $3.3\pm1.0$ &                                        $3.9\pm1.4$ &                                        $6.4\pm1.6$ &                                        $0.3\pm0.3$ &                                       $-0.4\pm0.5$ &                                       $-0.1\pm0.6$ \\
\addlinespace
\multicolumn{3}{l}{\textbf{Nesterov}}\\
                        0.3 &                      0.001 &                                        $1.5\pm0.7$ &                                        $1.5\pm0.6$ &                                        $2.9\pm1.1$ &                                        $3.5\pm0.8$ &                                        $5.3\pm1.4$ &                                        $6.5\pm1.1$ \\
                        0.3 &                       0.01 &                                        $4.5\pm1.1$ &                                        $5.0\pm1.2$ &                                        $8.5\pm1.5$ &                                        $2.6\pm0.8$ &                                        $3.0\pm1.2$ &                                        $3.5\pm1.5$ \\
                        0.3 &                        0.1 &                                        $3.3\pm1.0$ &                                        $3.4\pm1.0$ &                                        $6.5\pm1.5$ &                                        $0.5\pm0.3$ &                                        $0.1\pm0.7$ &                                       $-0.6\pm0.4$ \\ \addlinespace
                        0.6 &                      0.001 &                                        $1.6\pm0.8$ &                                        $1.9\pm0.7$ &                                        $4.0\pm1.3$ &                                        $4.0\pm0.9$ &                                        $5.6\pm1.4$ &                                        $6.7\pm1.4$ \\
                        0.6 &                       0.01 &                                        $3.4\pm1.1$ &                                        $4.3\pm1.0$ &                                        $6.9\pm1.4$ &                                        $0.9\pm0.3$ &                                        $1.2\pm0.7$ &                                        $1.4\pm0.7$ \\
                        0.6 &                        0.1 &                                        $2.8\pm1.0$ &                                        $2.8\pm1.0$ &                                        $5.5\pm1.2$ &                                        $0.4\pm0.4$ &                                       $-0.1\pm0.4$ &                                       $-0.2\pm0.6$ \\ \addlinespace
                        0.9 &                      0.001 &                                        $1.3\pm0.8$ &                                        $1.4\pm0.6$ &                                        $3.2\pm1.2$ &                                        $3.1\pm1.1$ &                                        $2.8\pm1.2$ &                                        $4.3\pm1.6$ \\
                        0.9 &                       0.01 &                                        $1.7\pm0.8$ &                                        $2.5\pm0.8$ &                                        $3.8\pm1.4$ &                                        $0.6\pm0.3$ &                                        $0.4\pm0.7$ &                                       $-0.8\pm0.4$ \\
                        0.9 &                        0.1 &                                        $1.7\pm0.8$ &                                        $1.8\pm0.8$ &                                        $3.3\pm1.1$ &                                        $0.3\pm0.2$ &                                        $0.4\pm0.8$ &                                       $-0.5\pm0.6$ \\
\addlinespace
\multicolumn{3}{l}{\textbf{Momentum}}\\
                        0.3 &                      0.001 &                                        $1.3\pm0.6$ &                                        $1.4\pm0.6$ &                                        $2.6\pm1.2$ &                                        $3.5\pm0.8$ &                                        $5.3\pm1.4$ &                                        $6.5\pm1.1$ \\
                        0.3 &                       0.01 &                                        $4.0\pm1.2$ &                                        $4.8\pm1.2$ &                                        $7.6\pm1.3$ &                                        $3.0\pm1.0$ &                                        $3.5\pm1.4$ &                                        $4.1\pm1.5$ \\
                        0.3 &                        0.1 &                                        $2.6\pm0.9$ &                                        $2.8\pm0.9$ &                                        $5.9\pm1.3$ &                                        $0.6\pm0.3$ &                                        $0.0\pm0.4$ &                                       $-0.7\pm0.4$ \\\addlinespace
                        0.6 &                      0.001 &                                        $1.5\pm0.7$ &                                        $1.4\pm0.7$ &                                        $3.5\pm1.2$ &                                        $4.2\pm1.1$ &                                        $6.1\pm1.6$ &                                        $6.9\pm1.5$ \\
                        0.6 &                       0.01 &                                        $3.0\pm1.1$ &                                        $4.0\pm1.0$ &                                        $5.3\pm1.2$ &                                        $1.3\pm0.4$ &                                        $1.4\pm1.0$ &                                        $1.0\pm0.6$ \\
                        0.6 &                        0.1 &                                        $2.1\pm0.9$ &                                        $2.2\pm0.9$ &                                        $4.4\pm1.1$ &                                        $0.4\pm0.4$ &                                        $0.6\pm0.6$ &                                       $-0.4\pm0.6$ \\ \addlinespace
                        0.9 &                      0.001 &                                        $0.9\pm0.7$ &                                        $0.7\pm0.7$ &                                        $2.4\pm1.2$ &                                        $2.2\pm0.7$ &                                        $1.3\pm0.7$ &                                        $2.1\pm0.8$ \\
                        0.9 &                       0.01 &                                        $1.7\pm0.8$ &                                        $2.2\pm1.1$ &                                        $2.8\pm1.0$ &                                        $0.7\pm0.3$ &                                        $0.3\pm0.5$ &                                       $-0.2\pm0.5$ \\
                        0.9 &                        0.1 &                                        $1.4\pm0.9$ &                                        $1.4\pm0.7$ &                                        $2.5\pm1.0$ &                                        $0.3\pm0.3$ &                                       $-0.3\pm0.6$ &                                       $-0.2\pm0.6$ \\
\bottomrule
\end{tabular}
}
\end{table}

\begin{table}[b]
\caption{\label{APDX:tab:momentum_head_classifier}
\textbf{Momentum for classifier or feature extractor only.} Considers Nesterov-momentum for learning rate $\eta=0.01$ and momentum strength $\rho$ for only the classifier ($\rho_{\text{head}}$) or feature extractor ($\rho_{\text{feat}}$).
Reports mean ($\pm \text{SE}$) over users in $\mathcal{U}_{\text{train}}$.
}
\centering
\resizebox{\linewidth}{!}{%
\begin{tabular}{rrllllll}
\toprule
 \multicolumn{1}{c}{$\rho_{\text{head}}$} &  \multicolumn{1}{c}{$\rho_{\text{feat}}$} & \multicolumn{1}{c}{$\overline{\text{OAG}}_{\text{action}}$} & \multicolumn{1}{c}{$\overline{\text{OAG}}_{\text{verb}}$} & \multicolumn{1}{c}{$\overline{\text{OAG}}_{\text{noun}}$} & \multicolumn{1}{c}{$\overline{\text{HAG}}_{\text{action}}$} & \multicolumn{1}{c}{$\overline{\text{HAG}}_{\text{verb}}$} & \multicolumn{1}{c}{$\overline{\text{HAG}}_{\text{noun}}$} \\
\midrule
                        0.0 &                       0.0 &                                        $4.9\pm1.2$ &                                        $5.5\pm1.6$ &                                        $8.9\pm1.5$ &                                        $2.6\pm0.8$ &                                        $3.6\pm1.2$ &                                        $4.8\pm1.7$ \\
\addlinespace
                                      0.0 &                                       0.3 &                                                 $4.8\pm1.2$ &                                               $5.4\pm1.5$ &                                               $8.5\pm1.4$ &                                                 $2.8\pm0.9$ &                                               $2.7\pm0.8$ &                                               $4.5\pm1.7$ \\
                                      0.0 &                                       0.6 &                                                 $4.3\pm1.1$ &                                               $5.0\pm1.4$ &                                               $7.9\pm1.4$ &                                                 $1.6\pm0.4$ &                                               $1.4\pm0.9$ &                                               $3.4\pm1.7$ \\
                                      0.0 &                                       0.9 &                                                 $3.7\pm1.2$ &                                               $4.4\pm1.3$ &                                               $6.7\pm1.4$ &                                                 $0.7\pm0.5$ &                                               $0.6\pm0.7$ &                                               $0.7\pm0.6$ \\ \addlinespace
                                      0.3 &                                       0.0 &                                                 $4.2\pm1.1$ &                                               $5.2\pm1.3$ &                                               $8.1\pm1.3$ &                                                 $3.1\pm0.9$ &                                               $4.0\pm1.5$ &                                               $4.8\pm1.4$ \\
                                      0.6 &                                       0.0 &                                                 $3.4\pm1.1$ &                                               $4.6\pm1.1$ &                                               $6.5\pm1.2$ &                                                 $3.6\pm0.9$ &                                               $5.1\pm1.7$ &                                               $5.6\pm1.7$ \\
                                      0.9 &                                       0.0 &                                                 $2.0\pm0.8$ &                                               $3.0\pm0.9$ &                                               $3.5\pm1.0$ &                                                 $1.1\pm0.3$ &                                               $1.1\pm0.5$ &                                               $1.1\pm0.6$ \\
\bottomrule
\end{tabular}
}
\end{table}

\subsection{Gradient analysis for online finetuning}
To investigate the inefficacy of momentum for SGD, we perform a gradient analysis in the following.
On top of the current batch gradient $g_t=\nabla_{\theta_t}\mathcal{L}_t$ at time step $t$, momentum adds a velocity gradient vector that is an exponentially moving average of the gradients in previous timesteps. The gradient vector of $k$ steps before $t$ hence diminishes in magnitude as $k$ increases, but retains its direction.  
Accelerating optimization for the current batch would require the gradients of current and previous time steps to have the same direction, resulting in a positive dot-product.
Additionally normalizing the gradient vectors, we report the cosine-similarity for $k\in[1,10]$ steps before $t$, averaged over all SGD updates per user stream, and equally weighed over users in $\mathcal{U}_{\text{train}}$.
Figure~\ref{fig:SGD_grad_analysis} shows near-zero gradient cosine similarity ($\cos_\angle$) for all $k$.
Noteably, the recent batch gradients have the largest variation, indicating either strong agreement or disagreement of gradient direction. The noisy gradient results indicate momentum's inefficacy on EgoAdapt.

Additionally, Table~\ref{APDX:tab:SGD_gradient_analysis_history} reports the numerical results and besides the cosine similarity for the full model ($F \circ H$), the video encoder ($F$) and classifier head ($H$) only, we also report cosine-similarity of sub-gradients for the Slow ($F_\text{slow}$) and Fast ($F_\text{fast}$) submodules.

\begin{table}[b]
\caption{\label{APDX:tab:SGD_gradient_analysis_history}
\textbf{Sub-gradient alignment analysis for finetuning} comparing the cosine-similarity of the current gradient for batch at time-step $t$ with gradient of $k$ time steps back (at time step $t-k$). Positive cosine-similarity implies constructive interference, whereas negative cosine-similarity results in a decrease of batch $t-k$'s loss when updating on batch $t$.
All results are first averaged for all ($t$, $t-k$) gradient pairs per user-stream, and subsequently averaged ($\pm SE$) over users in $\mathcal{U}_\text{train}$. 
We report the cosine-similarity of sub-gradients for the full SlowFast model ($F \circ H$), with feature extractor ($F$) consisting of a slow ($F_{\text{slow}}$) and fast  ($F_{\text{fast}}$) video encoder, and classifier head $H$.
}
\centering
\resizebox{\linewidth}{!}{%
\begin{tabular}{rrrrrr}
\toprule
 \multicolumn{1}{c}{$k$} & \multicolumn{1}{c}{$F \circ H$} & \multicolumn{1}{c}{$F_\text{slow}$} & \multicolumn{1}{c}{$F_\text{fast}$} & \multicolumn{1}{c}{$H$} & \multicolumn{1}{c}{$F$} \\
\midrule
                                 1 &                              $0.035\pm0.093$ &                             $-0.002\pm0.038$ &                              $0.014\pm0.034$ &                              $0.038\pm0.132$ &                             $-0.002\pm0.038$ \\
                                 2 &                             $-0.045\pm0.056$ &                             $-0.014\pm0.021$ &                             $-0.021\pm0.027$ &                             $-0.063\pm0.083$ &                              $-0.015\pm0.02$ \\
                                 3 &                              $0.037\pm0.048$ &                              $0.016\pm0.017$ &                              $0.024\pm0.024$ &                                $0.07\pm0.09$ &                              $0.016\pm0.017$ \\
                                 4 &                              $0.027\pm0.067$ &                              $0.022\pm0.018$ &                              $0.028\pm0.017$ &                              $0.008\pm0.103$ &                              $0.022\pm0.018$ \\
                                 5 &                              $0.053\pm0.045$ &                               $0.025\pm0.02$ &                              $0.019\pm0.022$ &                              $0.087\pm0.081$ &                               $0.025\pm0.02$ \\
                                 6 &                              $0.026\pm0.037$ &                             $-0.004\pm0.016$ &                              $0.016\pm0.013$ &                              $0.063\pm0.065$ &                             $-0.004\pm0.016$ \\
                                 7 &                             $-0.001\pm0.049$ &                             $-0.009\pm0.017$ &                              $-0.03\pm0.019$ &                             $-0.002\pm0.073$ &                              $-0.01\pm0.017$ \\
                                 8 &                              $0.028\pm0.042$ &                             $-0.001\pm0.017$ &                              $0.001\pm0.017$ &                              $0.056\pm0.071$ &                             $-0.001\pm0.016$ \\
                                 9 &                              $0.042\pm0.073$ &                               $0.004\pm0.03$ &                              $0.049\pm0.031$ &                              $0.048\pm0.105$ &                               $0.006\pm0.03$ \\
                                10 &                              $0.051\pm0.037$ &                                $0.02\pm0.01$ &                             $-0.006\pm0.018$ &                              $0.077\pm0.062$ &                              $0.019\pm0.011$ \\
\bottomrule
\end{tabular}
}
\end{table}

\subsection{Verb-classifier parameter analysis} 
\begin{figure}[t]
\caption{
\textbf{Linear verb-classifier analysis} comparing learning of the head $H$ only to the full model $F \circ H$.
Per user the final classifier weight and bias $L2$-norms are compared to the initial population model.
The per-user delta distribution is averaged over users in $\mathcal{U}_{\text{train}}$, shown with shaded $SE$. Decreases w.r.t. the population model are displayed as negative.
Verbs are ordered based on the average frequency distribution $P_{\text{label}}$ in the stream (shaded area). 
}
\label{APDX:fig:classifier_only_weight_biases_analysis:VERB}
\centering
        \begin{subfigure}{0.5\linewidth}
      \centering
        \caption{weight norm delta}%
        \includegraphics[clip,trim={0.2cm 0.2cm 0cm 0.2cm},width=1\textwidth]{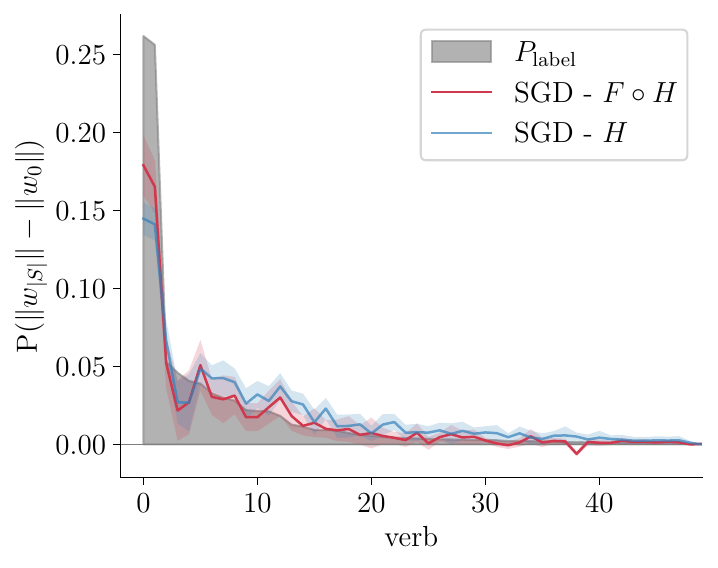} %
    \end{subfigure}%
        \begin{subfigure}{0.5\linewidth}
      \centering
              \caption{bias norm delta}%
        \includegraphics[clip,trim={0.2cm 0.2cm 0cm 0.2cm},width=1\textwidth]{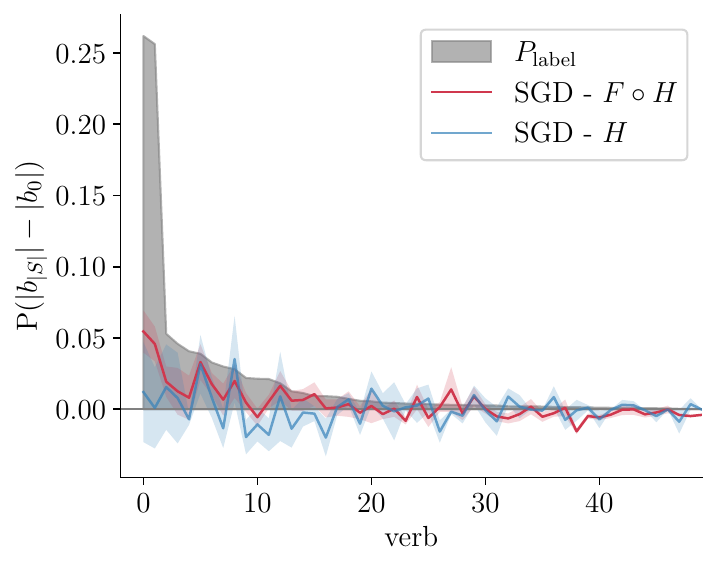} %
    \end{subfigure}%
\end{figure}

In the main paper, we show results for the noun-classifier's weight and bias L2-norms over the noun distribution. Additionally, Figure~\ref{APDX:fig:classifier_only_weight_biases_analysis:VERB} shows the results for the verb-classifier.
For the analysis, in both the verb and noun classifiers, both the weight and bias norms and the label distribution are calculated per user. Figure~\ref{APDX:fig:classifier_only_weight_biases_analysis:VERB} averages the per-user distributions and shows the means ($\pm SE$) over user distributions.

\subsection{Experience Replay and Hybrid-CBRS}
The Hybrid-CBRS storage strategy for ER is reported in Algorithm~\ref{algor:CBRS_adapted}. It combines the CBRS~\cite{chrysakis20a} and Reservoir~\cite{vitter1985random} methods by switching from CBRS to Reservoir sampling once the number of observed classes is greater than or equal to the memory size $M$.

ER results in the main paper perform an ablation on memory size $M$ and storage strategy with $\eta=0.01$. As only the action-based results are reported in the main paper, Table~\ref{APDX:tab:ER_full_results} reports the full results including verbs and nouns.
For the \textbf{linear probing} experiment, classifier retraining use batch size 32 and fixed learning rate $0.01$ for 10 epochs, considering best-performing Hybrid-CBRS with $M=64$ and SGD  both with learning rate $0.01$.

\begin{table}[t]
\caption{
\textbf{Experience Replay (ER) full results for actions, verbs, and nouns} for three  storage policies and memory sizes $M$.
Baseline \emph{ER-Full} stores all samples, and \emph{SGD} stores none.
Reported as mean ($\pm \text{SE}$) over users in $\mathcal{U}_{\text{train}}$. 
}
\label{APDX:tab:ER_full_results}
\centering
\resizebox{\linewidth}{!}{%
\begin{tabular}{rllllll}
\toprule
 $M$ & $\overline{\text{OAG}}_{\text{action}}$ & $\overline{\text{OAG}}_{\text{verb}}$ & $\overline{\text{OAG}}_{\text{noun}}$ & $\overline{\text{HAG}}_{\text{action}}$ & $\overline{\text{HAG}}_{\text{verb}}$ & $\overline{\text{HAG}}_{\text{noun}}$ \\
\midrule
\addlinespace\textbf{FIFO}\\
                                   8 &                                        $4.5\pm1.0$ &                                        $5.1\pm1.0$ &                                        $8.6\pm1.4$ &                                        $8.6\pm2.0$ &                                       $12.0\pm2.8$ &                                       $12.3\pm3.3$ \\
                                  64 &                                        $3.7\pm0.9$ &                                        $4.5\pm1.3$ &                                        $7.6\pm1.2$ &                                       $15.7\pm2.6$ &                                       $18.4\pm2.6$ &                                       $23.8\pm4.0$ \\
                                 128 &                                        $4.0\pm1.0$ &                                        $4.7\pm1.1$ &                                        $7.6\pm1.4$ &                                       $18.7\pm2.3$ &                                       $24.5\pm3.7$ &                                       $26.0\pm4.2$ \\
\addlinespace\textbf{Reservoir}\\
                                   8 &                                        $3.5\pm1.0$ &                                        $4.0\pm0.9$ &                                        $7.3\pm1.4$ &                                       $13.6\pm1.7$ &                                       $14.5\pm1.7$ &                                       $22.2\pm3.7$ \\
                                  64 &                                        $3.9\pm0.9$ &                                        $4.1\pm1.0$ &                                        $8.1\pm1.4$ &                                       $24.8\pm3.2$ &                                       $28.5\pm3.1$ &                                       $30.7\pm4.3$ \\
                                 128 &                                        $3.9\pm0.8$ &                                        $4.3\pm1.0$ &                                        $8.1\pm1.2$ &                                       $24.0\pm2.5$ &                                       $28.0\pm3.1$ &                                       $29.5\pm3.8$ \\
\textbf{Hybrid-CBRS}\\
                                   8 &                                        $3.9\pm1.0$ &                                        $4.4\pm1.0$ &                                        $7.9\pm1.5$ &                                       $15.6\pm2.5$ &                                       $19.5\pm3.1$ &                                       $21.6\pm3.7$ \\
                                  64 &                                        $4.6\pm0.9$ &                                        $4.9\pm1.0$ &                                        $8.9\pm1.4$ &                                       $29.7\pm4.7$ &                                       $34.0\pm4.5$ &                                       $37.0\pm5.4$ \\
                                 128 &                                        $4.1\pm0.9$ &                                        $4.8\pm0.9$ &                                        $8.7\pm1.4$ &                                       $25.1\pm4.1$ &                                       $26.3\pm5.3$ &                                       $38.5\pm4.4$ \\
\addlinespace\textbf{ER - full} &                                        $3.8\pm0.9$ &                                        $4.5\pm1.1$ &                                        $7.5\pm1.3$ &                                       $23.3\pm2.5$ &                                       $27.5\pm3.4$ &                                       $31.0\pm4.1$ \\

\bottomrule
\end{tabular}
}
\end{table}

\renewcommand{\algorithmicensure}{\textbf{Given:}}
\begin{algorithm}[t]
\begin{algorithmic}[1] %
\Ensure  replay memory $\mathcal{M}_c$ conditional on $c\in C$, a set of filled conditionals $\mathcal{F}$, total memory size $M$, sample $({\bf x}_t,{\bf y}_t)$ after observing $S_{0:t-1}$
\State $C \leftarrow C \cup \{{\bf y}_t\}$
\If {$|\mathcal{M}| < M$}
\State {Store $({\bf x}_t,{\bf y}_t)$ in $\mathcal{M}_{y_t}$ }
\ElsIf {$|C| \geq M$}
\State \textsc{Reservoir}$(\mathcal{M},({\bf x}_t,{\bf y}_t) )$ \Comment{Switch to reservoir sampling agnostic to conditionals}
\Else
\State $c^*=\argmax_{c} |\mathcal{M}_{c}|$, \  $\mathcal{F} \leftarrow \mathcal{F} \cup \{ c^*\}$
\If {${\bf y}_t \notin \mathcal{F}$} \Comment{Conditional memory not filled}
\State Remove random sample from $\mathcal{M}_{c^*}$
\State {Store $({\bf x}_t,{\bf y}_t)$ in $\mathcal{M}_{{\bf y}_t}$}
\Else %
\State \textsc{Reservoir}$(\mathcal{M}_{{\bf y}_t},({\bf x}_t,{\bf y}_t) )$ \Comment{Conditional reservoir sampling}
\EndIf
\EndIf
\end{algorithmic}
\caption{\label{algor:CBRS_adapted} Hybrid Class-balanced Reservoir Sampling}
\label{algor:framework}
\end{algorithm}

\subsection{User transfer study for verbs and nouns} 

\begin{figure*}[!ht]
\caption{\label{APDX:fig:transfer_eval:IOU}
\textbf{User labels intersection-over-union (IOU) } indicating the overlap of the action (a), verbs (b), and nouns (c) for the users in $\mathcal{U}_\text{train}$.
\looseness=-1
}
\centering
    \begin{subfigure}{0.32\linewidth}
      \centering
        \caption{$\text{IOU}_\text{action}(\%)$}%
        \includegraphics[clip,trim={0.2cm 0.2cm 2.8cm 0.6cm},width=1\textwidth]{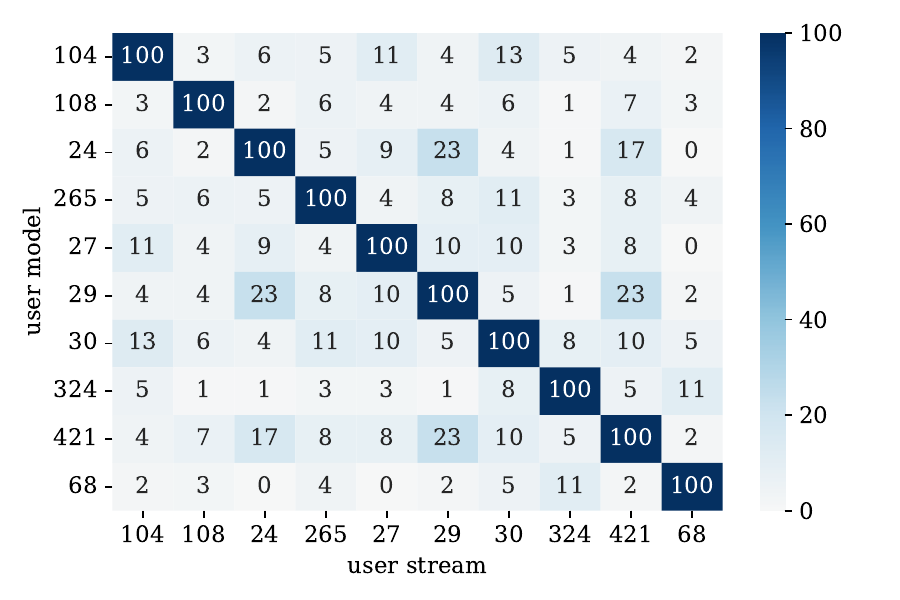} %
    \end{subfigure}%
        \begin{subfigure}{0.3\linewidth}
      \centering
        \caption{$\text{IOU}_\text{verb}(\%)$}%
        \includegraphics[clip,trim={0.8cm 0.2cm 2.8cm 0.6cm},width=1\textwidth]{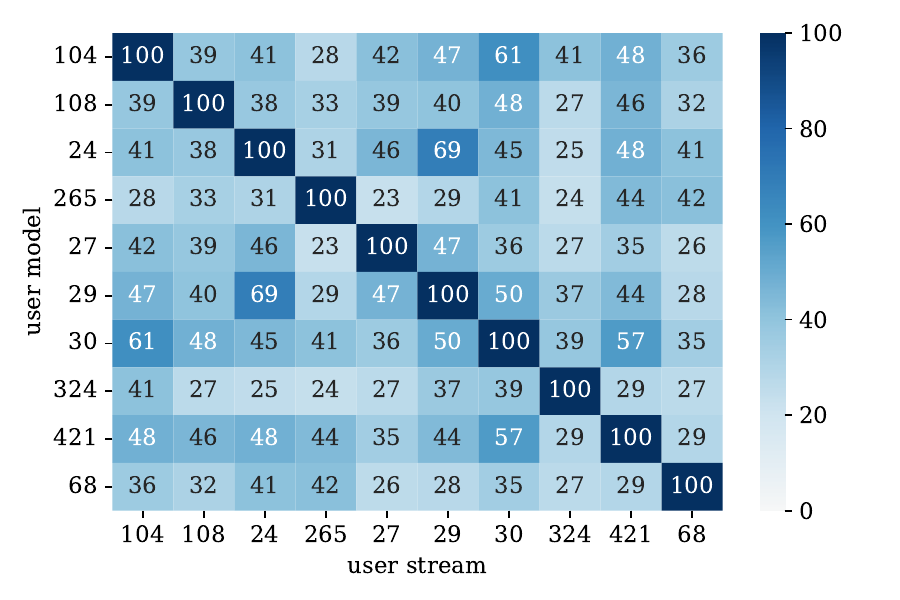} %
    \end{subfigure}%
        \begin{subfigure}{0.37\linewidth}
      \centering
        \caption{$\text{IOU}_\text{noun}(\%)$}%
        \includegraphics[clip,trim={0.8cm 0.2cm 0cm 0.2cm},width=1\textwidth]{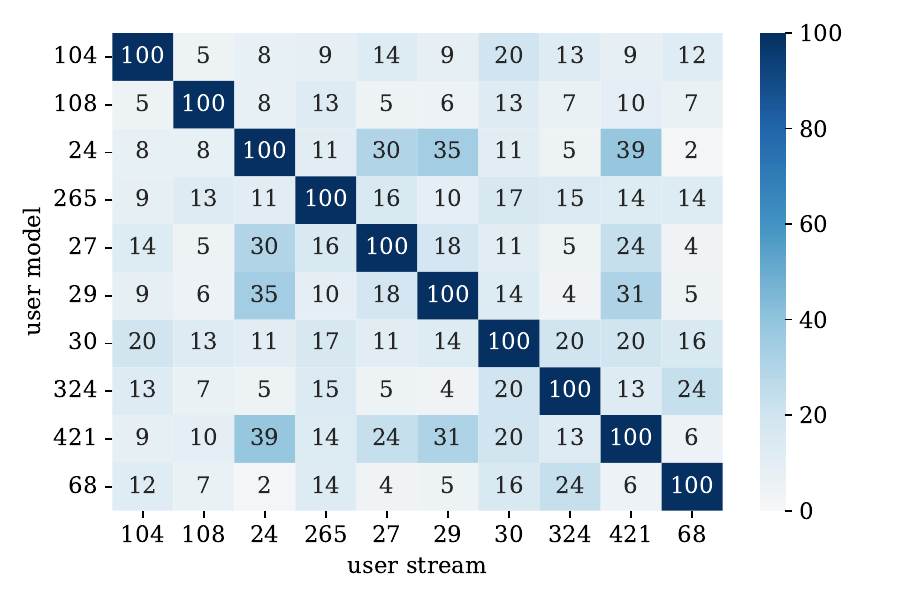} 
    \end{subfigure}%
\end{figure*}

\begin{figure}[!ht]
\caption{
\textbf{User transfer matrix} for users in $\mathcal{U}_{\text{train}}$ for verbs (a) and nouns (b). 
Rows represent user models $f_{\theta_{|S_u|}}$ after learning on user stream $S_u$.
Columns evaluate a row's user model on the various user streams.
Reports the loss in hindsight compared to the population model as $\text{HAG}_{\mathcal{L}}$.
}
\label{APDX:fig:transfer_eval:verb_noun}
\centering
        \begin{subfigure}{0.8\linewidth}
      \centering
          \caption{$\text{HAG}_{\mathcal{L},\text{verb}}$}
        \includegraphics[clip,trim={0.2cm 0cm 1.2cm 1.2cm},width=1\textwidth]{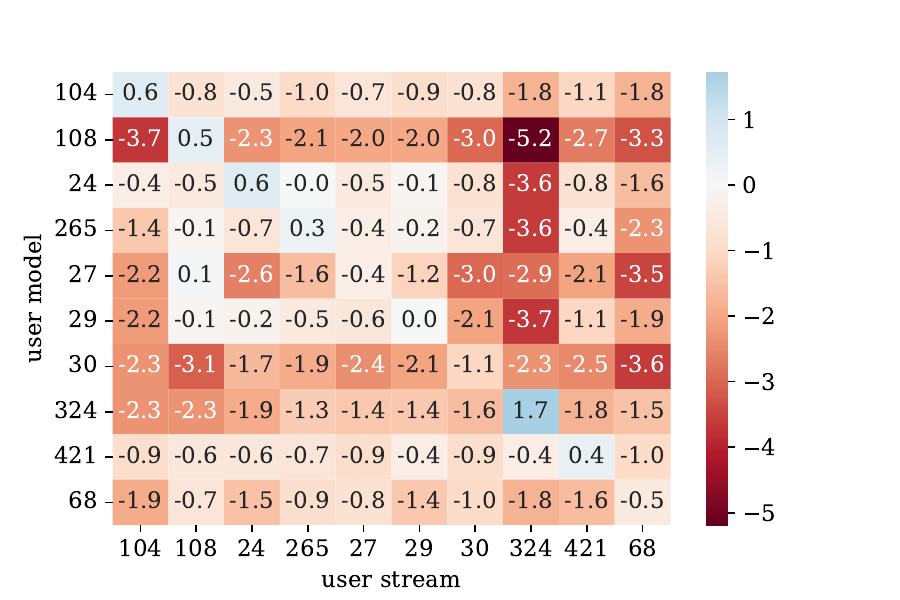} %
    \end{subfigure}
    \vspace{10pt} 
        \begin{subfigure}{0.8\linewidth}
      \centering
        \caption{$\text{HAG}_{\mathcal{L},\text{noun}}$}%
        \includegraphics[clip,trim={0.2cm 0.2cm 1.2cm 1.2cm},width=1\textwidth]{img/transfer_eval/micro_loss_based/noun_avg_AG_transfer_eval.pdf} 
    \end{subfigure}%

\end{figure}

Besides reporting the $\text{HAG}_{\mathcal{L},\text{action}}$ in the main paper, Figure~\ref{APDX:fig:transfer_eval:verb_noun} shows the user transfer matrices for $\text{HAG}_{\mathcal{L},\text{verb}}$ and $\text{HAG}_{\mathcal{L},\text{noun}}$. Similar to the action's user transfer matrix, we observe that the same general trend  persists for nouns on the diagonal, outperforming the population model. For verbs, it is more difficult to improve over the population model as user models 27,20,68 have negative adaptation gain in hindsight. This might be due to the high variability in the verbs, whereas the egocentric video is often concerned with only a single up to a few objects (or nouns) simultaneously.

To get an overview of the action overlap between users, Figure~\ref{APDX:fig:transfer_eval:IOU} reports the intersection-over-union (IOU) for the actions, verbs, and nouns between the users.
For example, users 24 and 29 are indicated to have similar actions, with $23\%$ overlap for the action domain, and $69\%$ for verbs,  $35\%$ for nouns.

\end{document}

%% file: tables/final_test_results_tabular_only.tex
\begin{tabular}{lllllll}
      & \multicolumn{3}{c}{online}                                                                      & \multicolumn{3}{c}{hindsight}                                                                   \\  \cmidrule(lr){2-4}\cmidrule(lr){5-7}
 \text{Method} & $\overline{\text{ACC}}_{\text{action}}$ & $\overline{\text{ACC}}_{\text{verb}}$ & $\overline{\text{ACC}}_{\text{noun}}$ & $\overline{\text{ACC}}_{\text{action}}$ & $\overline{\text{ACC}}_{\text{verb}}$ & $\overline{\text{ACC}}_{\text{noun}}$ \\
\midrule
\emph{Random}&$2.4e^{-3}$ & $0.9$ & $0.3$ & $2.4e^{-3}$ & $0.9$ & $0.3$ \\
\emph{Pretrain $\mathcal{U}_{\text{population}}$}&\NA{}&\NA{}&\NA{}& $1.2\pm0.2$ &                                        $5.9\pm0.5$ &                                        $4.2\pm0.4$\\
$\emph{LWP}_B$ & $7.7\pm0.4$ &                                   $14.9\pm0.9$ &                                   $18.2\pm1.0$ & $2.4\pm0.4$ &                                    $5.8\pm0.6$ &                                    $5.3\pm0.6$ \\
\midrule
\addlinespace\emph{1 update/batch}\\
\textbf{SGD} & $6.0\pm0.4$ &                                       $12.5\pm0.9$ &                                       $13.0\pm0.8$ &                                        $\bf 5.0\pm1.0$ &                                       $10.9\pm1.3$ &                                       $11.0\pm1.6$ \\
\textbf{SGD - head only} &                                        $\bf 7.0\pm0.5$ &                                       $14.3\pm0.6$ &                                       $16.9\pm0.9$ &                                        $2.8\pm0.4$ &                                        $8.0\pm1.5$ &                                        $6.6\pm0.7$ \\
\emph{SGD - i.i.d.} &                                        $5.5\pm0.5$ &                                       $11.3\pm1.0$ &                                       $13.8\pm1.2$ &                                       $15.1\pm1.3$ &                                       $22.9\pm1.6$ &                                       $28.8\pm1.9$ \\
\hdashline \addlinespace
\textbf{ER - $\textbf{FIFO}$}  &                                   $5.8\pm0.4$ &                                       $12.2\pm0.8$ &                                       $12.4\pm0.9$ &                                       $17.9\pm1.7$ &                                       $27.6\pm2.1$ &                                       $28.8\pm2.6$ \\
\textbf{ER - $\textbf{Reservoir}$}   &                                $5.9\pm0.5$ &                                       $12.1\pm0.8$ &                                       $12.4\pm0.9$ &                                       $24.9\pm1.5$ &                                       $34.3\pm2.0$ &                                       $37.4\pm2.2$ \\
\textbf{ER - Hybrid-CBRS} &                                        $5.9\pm0.5$ &                                       $12.1\pm0.8$ &                                       $12.8\pm0.9$ &                                       $\bf 34.2\pm1.8$ &                                       $40.1\pm2.2$ &                                       $48.0\pm2.4$ \\
\emph{ER - Full} &                                      $5.7\pm0.4$ &                                       $12.3\pm0.9$ &                                       $12.4\pm0.8$ &                                       $23.8\pm1.8$ &                                       $34.2\pm1.9$ &                                       $34.5\pm2.2$ \\

\midrule
\addlinespace\emph{10 updates/batch}\\
\textbf{SGD} &                                        $9.9\pm0.6$ &                                       $17.4\pm1.0$ &                                       $19.4\pm0.9$ &                                        $\bf 6.4\pm0.9$ &                                       $12.9\pm1.2$ &                                       $12.6\pm1.7$ \\
\textbf{SGD - head only} & $10.3\pm0.6$ &                                       $18.0\pm0.8$ &                                       $21.6\pm0.9$ &                                        $3.4\pm0.4$ &                                        $9.7\pm1.8$ &                                        $8.0\pm0.6$ \\
\emph{SGD - i.i.d.} &                                        $7.3\pm0.8$ &                                       $14.1\pm1.3$ &                                       $16.0\pm1.3$ &                                       $27.5\pm1.6$ &                                       $40.4\pm2.1$ &                                       $44.4\pm1.9$ \\ 
\hdashline \addlinespace
\textbf{ER - $\textbf{FIFO}$} & $10.6\pm0.6$ &                                       $18.4\pm1.1$ &                                       $19.9\pm0.9$ &                                       $53.6\pm3.4$ &                                       $62.2\pm3.1$ &                                       $59.2\pm3.6$ \\
\textbf{ER - $\textbf{Reservoir}$}& $10.5\pm0.6$ &                                       $18.0\pm0.9$ &                                       $19.4\pm0.9$ &                                       $58.6\pm3.1$ &                                       $66.7\pm3.0$ &                                       $65.5\pm2.9$ \\
\textbf{ER - Hybrid-CBRS} & $10.6\pm0.6$ &                                       $18.5\pm0.9$ &                                       $19.6\pm0.8$ &                                       $\bf 77.7\pm2.8$ &                                       $80.7\pm2.5$ &                                       $83.0\pm2.7$ \\
\emph{ER - Full} & $10.4\pm0.7$ &                                       $18.3\pm1.1$ &                                       $19.9\pm0.9$ &                                       $83.9\pm2.1$ &                                       $88.9\pm1.7$ &                                       $88.7\pm2.0$ \\
\bottomrule
\end{tabular}